\documentclass[11pt]{article}

\usepackage[preprint]{acl}

\usepackage{times}
\usepackage{latexsym}

\usepackage[T1]{fontenc}
\usepackage[utf8]{inputenc}
\usepackage{adjustbox}
\usepackage{microtype}

\usepackage{inconsolata}

\usepackage{graphicx}
\usepackage{xcolor}
\usepackage{tcolorbox}
\usepackage{algorithm}
\usepackage{algpseudocode}
\usepackage[normalem]{ulem}
\usepackage{multirow}
\usepackage{amsmath, amssymb, amsfonts}
\usepackage{booktabs}
\usepackage{float}
\usepackage{xspace} 
\usepackage{tabularx} 
\usepackage{enumerate}
\title{A Semiotics-Aware Framework for Evaluating Fidelity and Coverage in Natural Language Generation}

\author{
  Lorenzo Zangari \\
  University of Lausanne \\
  Lausanne, Switzerland \\
  \texttt{lorenzo.zangari@unil.ch}
  \And
  Davide Picca \\
  University of Lausanne \\
  Lausanne, Switzerland \\
  \texttt{davide.picca@unil.ch}
}

\newcommand{\meaningtext}{\textit{Meaning}}
\newcommand{\reference}{\textit{Reference}}
\newcommand{\expression}{\textit{Expression}}
\newcommand{\ifid}{\ensuremath{\operatorname{SF}}}
\newcommand{\icov}{\ensuremath{\operatorname{SC}}}
\newcommand{\fidelity}{Semiotic Fidelity}
\newcommand{\coverage}{Semiotic Coverage}
\newcommand{\wikipoly}{\textsc{WikiPoly}}
\newcommand{\commens}{\textsc{Commens}}
\newcommand{\peircelc}{\textsc{Peirce}}
\newcommand{\ose}{\textsc{OSE}}
\newcommand{\oseAE}{\textsc{OSE-AE}}
\newcommand{\oseAI}{\textsc{OSE-AI}}
\newcommand{\oseIE}{\textsc{OSE-IE}}
\newcommand{\swipe}{\textsc{SWiPE}}
\newcommand{\venus}{\emph{``Venus''}}
\newcommand{\refsec}[1]{Sec.~\ref{sec:#1}}

\newcommand{\pd}{\textsf{P}}
\newcommand{\qoned}{\textsf{Q1}}
\newcommand{\qtwod}{\textsf{Q2}}
\newcommand{\qthreed}{\textsf{Q3}}
\newcommand{\qfourd}{\textsf{Q4}}

\newcommand{\rone}{\text{R1}}
\newcommand{\rtwo}{\text{R2}}
\newcommand{\rthree}{\text{R3}}

\newcommand{\AlgComment}[1]{%
  \Statex\hspace{\algorithmicindent}%
  \parbox[t]{\dimexpr\linewidth-\algorithmicindent\relax}{%
    \footnotesize\color{gray!70!black}\texttt{//}~#1%
  }%
}

\begin{document}
\maketitle
\begin{abstract}
When two texts describe the same expression, standard metrics based on lexical overlap or whole-text similarity may fail to detect meaningful differences in how that expression is framed. We propose a framework to evaluate {semiotic alignment} between texts, where {a semiotic profile} encompasses both the contextual meaning and the discourse references made salient {by a text}. Our approach yields two scores, \emph{\fidelity{}} and \emph{\coverage{}}, estimating {how much of one text's profile is supported by the other and how much of the other's profile it recovers}. Experiments show that coverage is typically lower than fidelity, and that alignment between LLMs and human-curated data is highest at low sampling temperatures, while higher temperatures reduce this alignment. %
\end{abstract}
\section{Introduction}

\begin{figure}[!t]
    \centering
    \includegraphics[
        width=\linewidth%
    ]{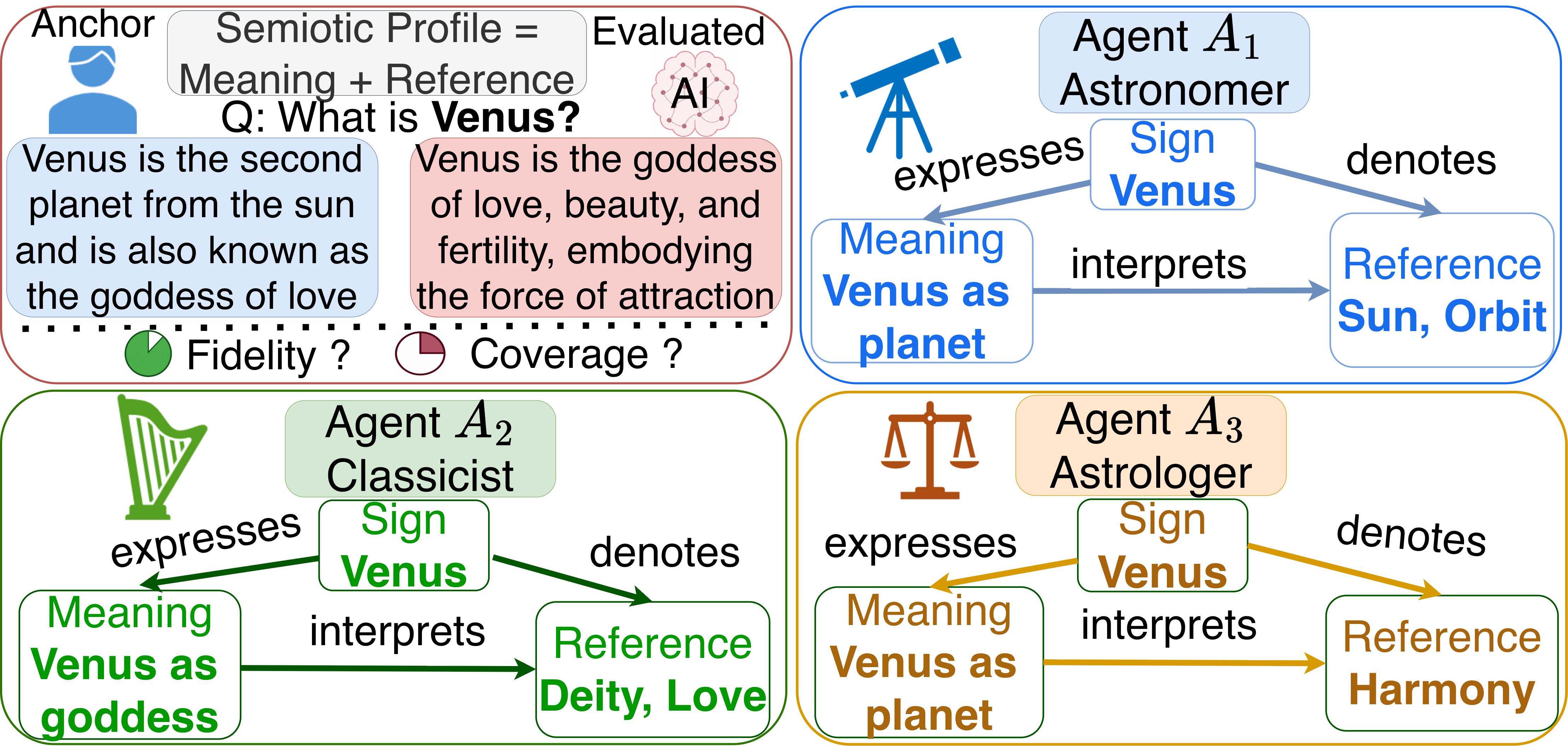}
\caption{Semiotic profiles of \venus: human vs. LLM (top left); semiotic triads for astronomer (top right), classicist (bottom left), astrologer (bottom right).}
    \label{fig:semiotic_triangle}
\end{figure}

The rapid rise of \emph{Large Language Models (LLMs)} has made natural language a central interface between humans and machines \citep{naveed2025comprehensive}, making the evaluation of \emph{Natural Language Generation (NLG)} increasingly important. The content 
of a linguistic expression, however, is often not fully determined by 
its surface form alone, as the same words can convey different 
meanings depending on who interprets them and on the context in 
which they occur \citep{haber-poesio-2024-polysemy,picca2025semioticchannelprinciplemeasuring}. As illustrated 
in Fig.~\ref{fig:semiotic_triangle}, the expression \venus{} may be 
described by one \emph{agent}
as the ``Roman goddess'', by another as ``the second planet from the Sun'', and by another as an astrological symbol, where \emph{we use the term agent to refer to any entity that produces text}. Even if such 
descriptions are fluent and coherent, they diverge both 
in the contextual meaning assigned to the expression and in the 
entities they bring into the discourse.  As humans reveal their communicative intent through  their linguistic choices \citep{levelt1993speaking}, LLMs commit  to a particular reading of the input in the act of generating text, 
where some properties become central, some entities become 
relevant, and other possible framings may be left unexpressed. Generated text can be treated as observable evidence of how the 
model has organized a sign into meaning 
\citep{lepori-etal-2025-racing}. This prompts us to address the following question in this work: \emph{given the 
same input, how can we estimate whether two agents produce overlapping output-level framings? 
}  

Existing NLG
metrics based on surface overlap or contextual similarity 
\citep{papineni-etal-2002-bleu,lin2004rouge,zhang2020bertscore} are 
useful for measuring textual quality, while distributional 
approaches compare sets of texts in a shared embedding space 
\citep{pillutla2021mauve,le-bronnec-etal-2024-exploring}. In both 
cases, however, the whole text is used as the comparison unit, so 
factors such as fluency, style, length, and topical content are 
conflated with how the input is {framed}, reflecting the broader 
gap between form and meaning in language understanding 
\citep{bender-koller-2020-climbing,trott-etal-2020-construing}.

We address this gap by proposing a framework grounded in semiotics \citep{peirce1931}, the discipline that studies how signs---ranging from single terms to entire documents \citep{picca2008lmm}---acquire meaning in context. 
Central to our framework is the notion of {a \emph{semiotic profile}, which we define as the contextual meaning assigned to a sign together with the discourse entities made salient through that meaning}. Given two texts describing the same expression, our framework compares the {semiotic profile expressed by} an \emph{evaluated agent} against that of an \emph{anchor agent}, taken as the reference point.
First, the output of each agent is encoded through a semiotically 
grounded representation that captures both the contextual meaning 
assigned to the target expression and the discourse entities 
associated with it, which we call a \emph{Semiotic Embedding}. To 
obtain it, we { begin by masking} a target expression (e.g., \venus) and its 
related discourse entities in the text and extract the hidden 
representations at the masked positions from a \emph{Masked 
Language Model (MLM)} \citep{modernbert}. {Then, we combine these representations  through a logarithmic map \citep{pennec2006intrinsic}}. %

{Finally}, we approximate the empirical support of each agent and compare the two manifolds through a support-based estimator \citep{kynkaanniemi2019improved}, 
yielding two scores, \emph{\fidelity{} (\ifid)}, measuring {how much of the evaluated profile is supported by the anchor}, and \emph{\coverage{} (\icov)}, measuring how much of the anchor's {profile} is recovered by the evaluated agent. 

We summarize our contributions as follows: 
\begin{itemize}
\item We propose a novel computational framework { that operationalizes Peircean semiotics} for evaluating differences in how texts frame an expression. 
By comparing the output of an evaluated agent with that of an anchor agent, 
we estimate both unsupported content and 
anchor-supported content that the evaluated agent fails to recover.

\item We introduce a model-agnostic procedure for computing \emph{Semiotic Embeddings}: semiotically grounded text encodings that provide proxy representations of how a text frames a linguistic expression and the relevant discourse entities made salient within that frame.

\item Our experiments across different LLMs, tasks, and datasets 
suggest that coverage is the main bottleneck relative to fidelity. We found that alignment depends on task structure and model sampling parameters: lower temperatures yielded the strongest alignment with human-curated data, whereas higher temperatures reduced this alignment and increased model-to-model agreement. Moreover, using the \emph{LLM-as-a-judge} paradigm, we employed three frontier LLMs as a proxy for human evaluation, finding positive correlations between their ratings and our scores.
\end{itemize}

\section{Background}\label{sec:preliminaries}

\paragraph{Agent.}
In this work, we use the term \textit{agent} for any human or 
model capable of engaging in communicative processes structured by 
semiotic mechanisms, that is, able to take an expression{, i.e., any linguistic unit,}  as input 
and generate a text {about it}.

\paragraph{Peircean Semiotics.}
Peircean semiotics studies how signs, i.e., anything that stands for  something else in some respect \citep{peirce1931}, become meaningful  for an interpreter. Unlike de Saussure's dyadic model, which defines 
the sign through the relation between \textit{signifier} and  \textit{signified} \citep{de1989cours}, Peirce's model 
includes three elements:
the \textit{representamen}, the material form of the sign, the  \textit{object}, what the sign stands for, and the \textit{interpretant}, the meaning that arises in the agent when the representamen is taken to stand for the object. We adopt this framework because it enables us to capture how different agents frame the same input. %

\paragraph{Semiotic modeling.}
We ground our framework in the computational approach of 
\citet{picca2008lmm}, in which Peirce's three elements are 
formalized as follows: the \textit{representamen} corresponds to 
the \expression, i.e., any linguistic unit such as a word, or an entire document;
the \textit{interpretant} corresponds to the \meaningtext, i.e., the 
contextual sense assigned to that expression by an agent; and the 
\textit{object} corresponds to the \reference, i.e., the entity 
or situation within the universe of discourse made accessible through 
interpretation. These three elements are connected by three directed relations, 
illustrated in Fig.~\ref{fig:semiotic_triangle}. The first relation, 
\textit{Expresses}, connects an \expression{} to its \meaningtext{}, 
reflecting how a linguistic unit conveys a particular sense in context. 
\textit{Denotes} connects an \expression{} to its \reference{}, encoding 
the fact that a linguistic unit points to some entity or situation in 
the universe of discourse. \textit{Interprets} connects a 
\meaningtext{} to a \reference{}, expressing the idea that it is only 
through meaning that a reference becomes identifiable.

Following this semiotic
model, \uline{we define the \textbf{semiotic profile} of an agent as a text-based encoding of its \emph{interpretation}: a situated process in which the agent assigns a context-dependent \meaningtext{} to an \expression{} and, through that \meaningtext{}, identifies the relevant \reference{}}.

\paragraph{Masked Word Embedding.}
We used \textit{masked word embedding} as a proxy for modeling
\meaningtext{} and \reference{} described above. As shown by \citet{yamada2021semantic}, 
masking disentangles frame-level structure from lexical identity, 
grouping instances by the frame they evoke, which comprises both 
the relational context that constitutes its sense and the semantic 
roles and entities it makes available. Masked word embeddings are contextual representations produced by an MLM at 
the position where a target word $w_i$ has been replaced with a 
mask token (e.g., \texttt{[MASK]}) within its context window 
$c_i$. Since MLMs are trained to recover a masked word from 
context,  the hidden state obtained by applying the MLM to the masked context $mask(c_i, w_i)$
reflects what the 
context expects at that position instead of the lexical identity 
of $w_i$ itself. %

\paragraph{Problem statement.}
Let $\mathcal{E}$ be a set of expressions, where each 
$e \in \mathcal{E}$ is a {target expression under analysis} (e.g., 
\venus). Given $e$, each agent $\gamma \in \{\beta, \alpha^{*}\}$ 
produces a collection of texts $\mathcal{D}_{e}^{(\gamma)}$ 
describing $e$, each $D \in \mathcal{D}_{e}^{(\gamma)}$ providing a 
distinct realization of the same subject. Our goal is to quantify 
{how the semiotic profile expressed by} an \emph{evaluated agent} 
$\beta$ differs from that of an \emph{anchor agent} $\alpha^{*}$.
For each agent $\gamma$, we derive from $\mathcal{D}_{e}^{(\gamma)}$ 
a set of representations $\mathcal{Z}^{(\gamma)}_{e}$, where each 
vector encodes a \meaningtext{} assigned by $\gamma$ to $e$ and the 
associated \reference{} made salient through that \meaningtext{}. 
We then define an evaluation function that, given 
$\mathcal{Z}^{(\beta)}_{e}$ and $\mathcal{Z}^{(\alpha^{*})}_{e}$, 
returns two complementary scores, namely \emph{\fidelity{}} 
($\ifid$), which quantifies how much of $\beta$'s {profile} is 
supported by $\alpha^{*}$, and \emph{\coverage{}} ($\icov$), which 
quantifies how much of $\alpha^{*}$'s {profile} is recovered 
by $\beta$.

\section{Related work}\label{sec:related_work}
The evaluation of NLG has traditionally relied on surface-form 
comparison. Lexical overlap metrics 
\citep{papineni-etal-2002-bleu, lin2004rouge} 
score generated outputs against aligned human references through 
$n$-gram agreement, while embedding-based metrics such as 
BERTScore \citep{zhang2020bertscore} leverage contextual 
similarity in a pretrained representation space.
Diversity-oriented metrics such as Distinct-$n$ 
\citep{li-etal-2016-diversity} and Self-BLEU 
\citep{zhu2018texygen} measure variation among generated texts, 
but their reliance on surface $n$-gram statistics provides only a 
coarse characterization of semantic diversity. 
\citet{giulianelli-etal-2023-comes} integrated lexical  %
and semantic variability to assess whether generator uncertainty 
is calibrated to human production variability.  Multi-task 
benchmark suites \citep{srivastava2023beyond} and LLM-as-a-judge 
protocols \citep{zheng2023judging} employ LLMs as automated 
evaluators to aggregate quality scores across tasks. %

Recent work in image generation showed that outputs of generative systems should be assessed along two dimensions, \emph{precision} and \emph{recall} \citep{sajjadi2018assessing}. A non-parametric, manifold-based formulation was proposed by \citet{kynkaanniemi2019improved}, projecting real and generated samples into a shared embedding space and defining precision as the fraction of generated samples falling within the support of the real ones, and recall as the fraction of real samples covered by the generated ones.
A similar distribution-based perspective recently emerged in 
\emph{open-ended text generation}, where many distinct continuations 
may be equally valid and no single reference adequately represents 
the space of acceptable outputs. MAUVE \citep{pillutla2021mauve} 
compared machine-generated texts with a human-text distribution by 
embedding texts with GPT-2 \citep{radford2019language} and computing 
a divergence frontier summarized as a single score, conflating 
quality and diversity. \citet{le-bronnec-etal-2024-exploring} 
extended the approach of \citet{kynkaanniemi2019improved} to the 
text domain, estimating the supports of human and generated texts 
in an embedding space to recover precision and recall as two 
distinct scores.

\paragraph{Comparison with prior work.}
\begin{figure*}
    \centering
    \includegraphics[width=1.0\linewidth]{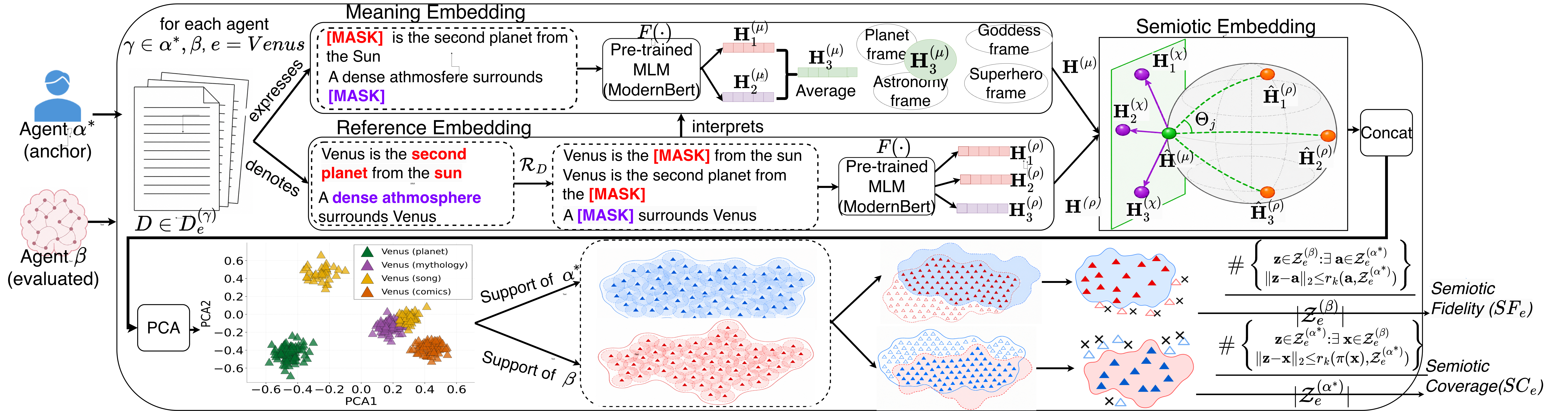}
    \caption{Our proposed framework to compute \fidelity{} (\ifid) and \coverage{} (\icov).}
    \label{fig:framework}
\end{figure*}

{\citet{picca2025semioticchannelprinciplemeasuring} employs the Peircean triad conceptually in an information-theoretic model, whereas we operationalize it as a computational framework.}
Our work is closest to \citet{le-bronnec-etal-2024-exploring} and 
\citet{pillutla2021mauve}, but diverges in its object of 
comparison. These distributional approaches embed each text as a 
single point, conflating heterogeneous factors such as fluency, 
style
and topical content into a global notion of quality 
and diversity \cite{tevet-berant-2021-evaluating}. Grounded in Peircean semiotics, our framework 
instead evaluates the \meaningtext{}--\reference{} configuration of 
a target expression, i.e., the contextual sense assigned to the topic 
and the discourse referents made salient through it. As shown in Fig.~\ref{fig:motivation}, this identifies fluent outputs reducing a polysemous topic to a small number of senses, and paraphrastic outputs preserving the same configuration under different surface forms.

\section{Methodology}
\paragraph{Overview.} Figure~\ref{fig:framework} shows our framework, grounded in 
semiotic principles, which compares two agents 
$\gamma \in \{\alpha^{*}, \beta\}$. First, we employ a Transformer 
encoder pre-trained with a masked language modeling objective to 
construct a \emph{Meaning Embedding}, capturing the contextual 
frame assigned to $e$ in each $D \in \mathcal{D}_{e}^{(\gamma)}$, 
and a \emph{Reference Embedding}, capturing the discourse entities salient in $D$. %
{ These two representations are then combined to encode each Reference as a point relative to the corresponding Meaning, following the semiotic model in \refsec{preliminaries}, yielding what we term \emph{Semiotic Embeddings}.} Then, we compare the Semiotic Embeddings of $\beta$ against 
those of $\alpha^{*}$. Specifically, for each agent, we estimate a 
manifold by approximating its support with hyperspheres whose radii 
are defined by local $k$-nearest-neighbor distances. \ifid{} and 
\icov{} are then computed as overlaps between the two empirical 
supports to measure whether $\beta$'s points remain within the 
anchor support and whether it covers the anchor, respectively.

\paragraph{Meaning Embedding.}
Given an \expression{} $e$ and a text $D \in 
\mathcal{D}_{e}^{(\gamma)}$ produced by an agent $\gamma$, we model 
the \meaningtext{} of $e$ as the semantic frame it evokes in $D$, 
which can be captured through masked word embeddings 
\citep{yamada2021semantic,zhou-etal-2019-bert}. To this end, we 
employ a Transformer encoder pre-trained with a masked language 
modeling objective, which we denote by $F$. Let $w_1,\ldots,w_n$ 
be the mentions of $e$ in $D$, including inflectional variants 
of $e$. For each mention $w_i$, we 
extract a context window $c_i$ consisting of the sentence 
containing $w_i$ and its left and right context. We then 
compute: %

\begin{equation}\label{eq:meaning_eq}
\scalebox{1.0}{$\displaystyle
\mathbf{H}^{(\mu)}_{i} = F\!\left(\mathrm{mask}(c_i, w_i)\right),
$}
\end{equation}

\noindent
where $\mathrm{mask}(c_i, w_i)$ denotes the context window $c_i$ 
with $w_i$ replaced by the special token of $F$, and 
$\mathbf{H}^{(\mu)}_{i}$ is the hidden representation of the 
special token at the last layer of $F$.\footnote{For multi-token occurrences of \(w_i\), the corresponding token representations are averaged.} The Meaning Embedding 
$\mathbf{H}^{(\mu)}$ is then obtained by averaging across all 
occurrences of $e$ in $D$:
\begin{equation}
\scalebox{1.0}{$\displaystyle
\mathbf{H}^{(\mu)}
=
\frac{1}{n}
\sum_{i=1}^{n}
\mathbf{H}^{(\mu)}_{i}.$}
\end{equation}
$\mathbf{H}^{(\mu)} \in \mathbb{R}^d$, with $d$ denoting the hidden 
dimensionality of $F$, serves as a proxy for the semantic frame  
{assigned to $e$ in $D$}, abstracting away from its 
lexical form. %

\paragraph{Reference Embedding.}
Given an \expression{} $e$ and a text $D \in 
\mathcal{D}_{e}^{(\gamma)}$ produced by an agent $\gamma$, we model 
its \reference{} as the entities made salient in $D$ by the frame 
evoked by $e$. Since \reference{} is frame-dependent, the same 
entity may play different roles across frames, and different 
surface forms may realize the same role. We therefore encode each 
\reference{} through masked word embeddings, reducing dependence on 
lexical identity while preserving its local context. Using the 
procedure in Appendix~\ref{app:srl}, we extract from $D$ the set 
$\mathcal{R}_D=\{r_1,\ldots,r_m\}$ of \reference{}s associated with 
$e$ (e.g., \emph{Sun} when $e=$\venus), excluding mentions of $e$ 
itself, already captured by the \meaningtext{}. For each $r_j$, let 
$w_{j,1},\ldots,w_{j,n_j}$ be its occurrences in $D$ and $c_{j,l}$ 
the context window around $w_{j,l}$. The Reference Embedding is 
obtained by averaging the masked representations over these 
occurrences:
\begin{equation}
\scalebox{1.0}{$\displaystyle
\mathbf{H}^{(\rho)}_{j}
=
\frac{1}{n_j}
\sum_{l=1}^{n_j}
F\!\left(\mathrm{mask}(c_{j,l}, w_{j,l})\right),
$}
\end{equation}

\noindent
where $\mathbf{H}^{(\rho)}_{j}$ is the average hidden representation 
of the masked occurrences at the last layer of $F$, the same MLM used 
for the Meaning Embedding, providing a contextual representation of 
$r_j$ that reduces dependence on its surface form.

\paragraph{Semiotic Embedding.}
The Semiotic Embedding encodes how an agent frames an \expression{} $e$ 
in a text $D \in \mathcal{D}_{e}^{(\gamma)}$ by representing each Reference Embedding 
relative to the Meaning Embedding, rather than as an independent 
point in the embedding space. We first project both the Meaning 
Embedding $\mathbf{H}^{(\mu)}$ and each Reference Embedding 
$\mathbf{H}^{(\rho)}_{j}$ onto the unit hypersphere by Euclidean 
normalization, yielding $\widehat{\mathbf{H}}^{(\mu)}$ and 
$\widehat{\mathbf{H}}^{(\rho)}_{j}$, respectively. We then apply 
the logarithmic map at 
$\widehat{\mathbf{H}}^{(\mu)}$, which expresses each normalized 
Reference Embedding through the direction and length of the 
shortest path along the hypersphere that connects it to the 
normalized Meaning Embedding. Let 
$\theta_j = \operatorname{\arccos}\!\left(\langle 
\widehat{\mathbf{H}}^{(\mu)}, \widehat{\mathbf{H}}^{(\rho)}_{j} 
\rangle \right)$ denote the angle between the two normalized 
embeddings, with $\langle\cdot,\cdot\rangle$ the Euclidean inner 
product. The spherical logarithmic map has the closed form:

{
\begin{equation}\label{eq:semiotic_displacement}
\scalebox{1.0}{$\displaystyle
\mathbf{H}^{(\chi)}_{j}
=
\theta_j
\frac{
\widehat{\mathbf{H}}^{(\rho)}_{j}
-
\cos(\theta_j)\,\widehat{\mathbf{H}}^{(\mu)}
}{
\left\|
\widehat{\mathbf{H}}^{(\rho)}_{j}
-
\cos(\theta_j)\,\widehat{\mathbf{H}}^{(\mu)}
\right\|
},
$}
\end{equation}}

\noindent
where $\|\cdot\|$ denotes the Euclidean norm, and 
$\mathbf{H}^{(\chi)}_{j}$ is the displacement that encodes each 
Reference Embedding relative to the Meaning Embedding. Then, we 
define the Semiotic Embedding as:

{
\begin{equation}
\scalebox{1.0}{$\displaystyle
\mathbf{H}^{(\sigma)}_{j}
=
\begin{cases}
\left[
\widehat{\mathbf{H}}^{(\mu)}
\,\Vert\,
\mathbf{0}_d
\right],
& j=0,\\
\left[
\widehat{\mathbf{H}}^{(\mu)}
\,\Vert\,
\mathbf{H}^{(\chi)}_{j}
\right],
& j=1,\ldots,m,
\end{cases}
$}
\end{equation}
}
\noindent
where $[\,\cdot\,\Vert\,\cdot\,]$ denotes vector concatenation and 
$\mathbf{H}^{(\sigma)}_{j} \in \mathbb{R}^{2d}$. The first 
component represents the \meaningtext{} assigned to $e$ in $D$, 
while the second represents which reference is reached from that 
\meaningtext{} and along which displacement. As a consequence, 
identical displacements attached to different \meaningtext{}s 
remain distinct through the first component. The case $j=0$ stores 
only the Meaning Embedding, independently of any Reference 
Embedding. %

\begin{figure*}[t!]
    \centering
    \includegraphics[
        width=\linewidth
    ]{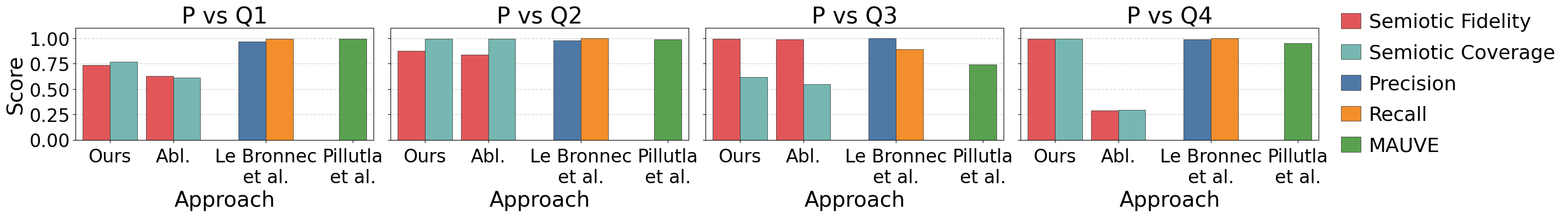}
    \caption{%
Controlled diagnostic comparing our framework and its log-map 
ablation with prior work.
    }
    \label{fig:motivation}
\end{figure*}

\paragraph{Semiotic Fidelity and Coverage.}
For a given \expression{} $e$, each agent $\gamma \in 
\{\beta, \alpha^{*}\}$ produces its own set of documents 
$\mathcal{D}_{e}^{(\gamma)}$ describing $e$, from which 
we compute its Semiotic Embeddings. We compare the evaluated agent 
$\beta$ against the anchor $\alpha^{*}$ through a manifold estimated 
from these embeddings. Since this  estimation relies on local distances, we first apply PCA
to the Semiotic Embeddings, retaining the smallest number of components
that explain at least $90\%$ of the variance, to reduce dimensionality
and filter noise before support estimation
\citep{le-bronnec-etal-2024-exploring}.
We 
denote by $\widetilde{\mathbf{H}}^{(\sigma)}_{D,j}$ the projected 
Semiotic Embedding for $\gamma$. For each agent, we then collect these projected 
representations across all documents $D \in \mathcal{D}_{e}^{(\gamma)}$ 
into a collection $\mathcal{Z}^{(\gamma)}_{e} = \bigcup_{D 
\in \mathcal{D}_{e}^{(\gamma)}} 
\{\widetilde{\mathbf{H}}^{(\sigma)}_{D,j}\}_{j=0}^{m_{\gamma,D}}$, 
where $m_{\gamma,D}$ is the number of references extracted from 
document $D$. We then approximate the support of 
$\mathcal{Z}^{(\gamma)}_{e}$ by centering a hypersphere on each 
semiotic point, with radius given by the distance to its $k$-th 
nearest neighbor within the full collection 
$\mathcal{Z}^{(\gamma)}_{e}$, computed across all documents of agent 
$\gamma$. 
Denoting by 
$\mathbf{N}_k(\widetilde{\mathbf{H}}^{(\sigma)}_{D,j}, 
\mathcal{Z}^{(\gamma)}_{e})$ the $k$-th nearest neighbor of 
$\widetilde{\mathbf{H}}^{(\sigma)}_{D,j}$ in 
$\mathcal{Z}^{(\gamma)}_{e} \setminus 
\{\widetilde{\mathbf{H}}^{(\sigma)}_{D,j}\}$, the radius is:

{
\begin{equation}
\scalebox{1.0}{$\displaystyle r_k\!\left(\widetilde{\mathbf{H}}^{(\sigma)}_{D,j}, 
\mathcal{Z}^{(\gamma)}_{e}\right)
=
\left\|
\widetilde{\mathbf{H}}^{(\sigma)}_{D,j}
-
\mathbf{N}_k\!\left(\widetilde{\mathbf{H}}^{(\sigma)}_{D,j}, 
\mathcal{Z}^{(\gamma)}_{e}\right)
\right\|_{2}.$}
\end{equation}
}

Given $\mathbf{x} \in \mathcal{Z}^{(\beta)}_{e}$ and 
$\mathbf{a} \in \mathcal{Z}^{(\alpha^{*})}_{e}$, \fidelity{} for 
\expression{} $e$ is defined as the fraction of evaluated points 
that fall within at least one hypersphere of the anchor support:

{
\begin{equation}
\scalebox{1.0}{$\displaystyle
\begin{aligned}
\ifid_e(\beta \mid \alpha^{*})
&=
\frac{1}{|\mathcal{Z}^{(\beta)}_{e}|}
\sum_{\mathbf{x} \in \mathcal{Z}^{(\beta)}_{e}}
\mathbb{I}\!\left[
\exists \mathbf{a} \in \mathcal{Z}^{(\alpha^{*})}_{e}:
\right.
\\[-0.2em]
&\left.
\|\mathbf{x} - \mathbf{a}\|_2
\le
r_k\!\left(\mathbf{a}, \mathcal{Z}^{(\alpha^{*})}_{e}\right)
\right].
\end{aligned}
$}
\end{equation}
}

For \coverage{}, a symmetric definition that estimates the $\beta$ 
support using radii $r_k(\mathbf{x}, \mathcal{Z}^{(\beta)}_{e})$ 
would be less suitable in our setting. Since LLMs can be unstable and sometimes produce outputs with few or sparse references, the 
$k$-nearest-neighbor distances internal to 
$\mathcal{Z}^{(\beta)}_{e}$ may become large, so that the 
$\beta$ support could expand when the evaluated agent produces less {referential content} \citep{pmlr-v119-naeem20a,Park2023ProbabilisticPA}. We therefore estimate the $\beta$ support at 
the local resolution of the anchor. For each point $\mathbf{x} \in 
\mathcal{Z}^{(\beta)}_{e}$, we center a hypersphere on $\mathbf{x}$ 
with the radius of its nearest anchor point, so that the radii 
reflect the density of the anchor rather than that of the 
evaluated agent. Following \citet{pmlr-v119-naeem20a}, \coverage{} is then the fraction of anchor points 
covered by at least one such hypersphere:
\begin{equation}
\label{eq:semiotic_fidelity_coverage}
\scalebox{1.0}{$\displaystyle
\begin{aligned}
\icov_e(\beta \mid \alpha^{*})
&=
\frac{1}{|\mathcal{Z}^{(\alpha^{*})}_{e}|}
\sum_{\mathbf{a} \in \mathcal{Z}^{(\alpha^{*})}_{e}}
\mathbb{I}\!\left[
\exists \mathbf{x} \in \mathcal{Z}^{(\beta)}_{e}:
\right.
\\[-0.2em]
&\left.
\|\mathbf{a} - \mathbf{x}\|_2
\le
r_k\!\left(
\pi(\mathbf{x}),
\mathcal{Z}^{(\alpha^{*})}_{e}
\right)
\right],
\end{aligned}
$}
\end{equation}

\noindent
where $\pi(\mathbf{x}) = \arg\min_{\mathbf{a} \in 
\mathcal{Z}^{(\alpha^{*})}_{e}} \|\mathbf{x} - \mathbf{a}\|_2$ denotes 
the nearest anchor point to $\mathbf{x}$. Unlike the formulation of 
\citet{le-bronnec-etal-2024-exploring}, this anchor-calibrated 
construction satisfies a monotonicity property, since the radius of 
each hypersphere is independent of the other evaluated points. %
Dropping a point from $\mathcal{Z}^{(\beta)}_{e}$ removes its 
hypersphere while leaving the others unchanged, so that removing 
points cannot increase \coverage{}.%

The final scores are computed by 
averaging the per-expression scores over all expressions 
$e \in \mathcal{E}$:
\begin{equation}
\label{eq:final_fidelity_coverage}
\scalebox{1.0}{$\displaystyle
\begin{aligned}
\ifid(\beta \mid \alpha^{*})
&=
\frac{1}{|\mathcal{E}|}
\sum_{e\in\mathcal{E}}
\ifid_e(\beta \mid \alpha^{*}),
\\
\icov(\beta \mid \alpha^{*})
&=
\frac{1}{|\mathcal{E}|}
\sum_{e\in\mathcal{E}}
\icov_e(\beta \mid \alpha^{*}).
\end{aligned}$}
\end{equation}
Thus, \ifid{} quantifies the extent to which {$\beta$'s profile is supported by $\alpha^{*}$}, 
while \icov{} quantifies the extent to which $\beta$ recovers $\alpha^{*}$'s profile.

\section{Experiments}

 \paragraph{Setup.}
{To evaluate whether \ifid{} and \icov{} follow the expected 
behavior, we 
first conduct a controlled diagnostic. We then evaluate the framework, 
which can compare the semiotic profiles of any pair of agents, across 
three settings.} In the \textbf{\rone{}} setting, the anchor is a human and the evaluated agents are LLMs. In \textbf{\rtwo{}}, both the anchor and the evaluated agents are LLMs. In \textbf{\rthree{}}, both the anchor and the evaluated agents are humans. We used $10$ random seeds to account for stochasticity and 
report average \ifid{} and \icov{}. For \textbf{\rone{}} and \textbf{\rtwo{}}, which involve LLM generation, we additionally varied the temperature over the range $[0.2, 1.6]$ in increments of $0.2$ to assess how generation stability affects {semiotic alignment}.
{Following \citet{kynkaanniemi2019improved}, we set $k=3$ 
in all experiments to retain a local support estimate and limit the 
support expansion induced by larger neighborhoods.} 
\paragraph{Data.}
For \textbf{\rone{}}, we employed \wikipoly, a dataset of 
lead-section paragraphs from Wikipedia pages associated with 
polysemous terms. For \textbf{\rtwo{}}, we additionally evaluated 
the ability of LLMs to generate content related to Peircean 
concepts, i.e., abstract theoretical terms derived from Peirce's 
semiotic theory, using \commens{}, a dataset extracted from the 
Commens project \citep{commens}. For \textbf{\rthree{}}, we 
employed two document-level simplification datasets, \swipe{} 
\citep{swipe} and \ose{} \citep{vajjala-lucic-2018-onestopenglish}, 
in which each complex text is paired with a human-written 
simplified version. \ose{} further provides three reading levels, 
Advanced, Intermediate, and Elementary, which we evaluate in the 
\oseAI{}, \oseAE{}, and \oseIE{} settings. We additionally built a dataset called \peircelc{} from Commens, splitting Peirce's 
writings into two temporally balanced halves, 
using the earlier half as anchor texts.

\paragraph{Models.}
\begin{table}[t]
    \centering
    \begingroup
    \footnotesize
    \renewcommand{\arraystretch}{0.55}
    \setlength{\aboverulesep}{0.1pt}
    \setlength{\belowrulesep}{0.1pt}
    \setlength{\extrarowheight}{0.1pt}
    \scalebox{0.95}{%
    \begin{tabular}{l|llc}
        \toprule
        \textbf{From} & \textbf{Model} & \textbf{Abbrev.} & \textbf{Params} \\
        \midrule
        \multirow{3}{*}{US}
        & \texttt{Phi-4-mini-instruct} & Phi-4 & 3.80B \\
        & \texttt{gemma-3-12b-it} & Gemma & 12.19B \\
        & \texttt{gpt-oss-20b} & \texttt{GPT-oss} & 20.9B \\
        \midrule
        EU & \texttt{Ministral-3-14B-Instruct} & Ministral & 13.90B \\
        \midrule
        \multirow{2}{*}{China}
        & \texttt{Qwen3-4B-Instruct} & Qwen & 4.00B \\
        & \texttt{glm-4-9b-chat} & GLM & 9.00B \\
        \midrule
        World & \texttt{tiny-aya-global} & Tiny-Aya & 3.35B \\
        \bottomrule
    \end{tabular}%
    }
    \endgroup
    \caption{LLMs selected for our study, annotated with their geographic ``location'' and number of parameters.}
    \label{tab:models}
\end{table}

We employed a %
selection of \emph{open LLMs} shown in Tab. \ref{tab:models}, with different sizes and architectures, using publicly available implementations %
on 
Hugging Face. %
As the pre-trained Transformer $F$, we used ModernBERT
\citep{modernbert}. %

{
\subsection{Controlled Diagnostic}
\label{sec:controlled_diagnostic}
{We test whether \ifid{} and \icov{} exhibit the expected 
behavior under controlled changes in sense coverage and surface form. 
We use \pd{}, an anchor dataset derived from \wikipoly{} in which each 
target expression has four lead-section paragraphs covering three 
senses, with one represented by two distinct paragraphs. For 
each instance in \pd{}, we construct four variants. \qoned{} replaces 
one paragraph with another from the same sense and a second with one 
from an unseen sense. We expect both scores to be near $0.75$, since 
three of the four components in \qoned{} are supported by \pd{} and 
three of those in \pd{} are recovered by \qoned{}. \qtwod{} retains 
\pd{} and adds paragraphs expressing additional senses, so \icov{} 
should remain near $1.00$ while \ifid{} decreases. \qthreed{} retains 
only a subset of \pd{}, so \ifid{} should remain near $1.00$ while 
\icov{} decreases. \qfourd{} replaces each paragraph with a paraphrase 
preserving its meaning and referents, so both scores should remain near 
$1.00$.

Figure~\ref{fig:motivation} compares our method with those of 
\citet{le-bronnec-etal-2024-exploring} and 
\citet{pillutla2021mauve}, and with an ablation of our framework 
without the logarithmic-map Reference displacement (Abl.). \ifid{} and \icov{} 
follow the expected patterns across all four transformations. On 
\qoned{}, our method yields $0.74/0.77$, reflecting unsupported and 
missing anchor content, while the prior methods provide high scores. 
Their scores also remain high on \qtwod{}, where our scores show the 
expected asymmetry between reduced \ifid{} and high \icov{}. 
On \qthreed{}, high \ifid{} and lower \icov{} capture the subset 
relation. Precision and recall reflect it more weakly, while 
MAUVE does not distinguish missing from unsupported content. On 
\qfourd{}, all methods except the ablation retain high scores under 
paraphrase. The ablation yields $0.61/0.58$ on \qoned{} and scores 
\qfourd{} even lower despite its preservation of all senses and 
referents. %

}

}

\subsection{\rone: LLMs vs. Human-curated data}

\begin{table}[!t]
\centering
\begingroup
\scriptsize
\setlength{\tabcolsep}{1pt}
\renewcommand{\arraystretch}{0.55}
\resizebox{1.01\columnwidth}{!}{%
\begin{tabular}{@{}lccccccc@{}}
\toprule
T & GPT-oss & Ministral & Gemma & GLM & Qwen & Phi-4 & Tiny-Aya \\
\specialrule{\lightrulewidth}{0pt}{0.15ex}
0.2 & $.16/.19$ & $.16/.22$ & $\underline{\mathbf{.13}/.22}$ & $.18/.14$ & $\underline{\mathbf{.19}/.25}$ & $.28/.20$ & $.25/.16$ \\
0.4 & $.16/.21$ & $\underline{.15/\mathbf{.23}}$ & $.12/.21$ & $.16/.14$ & $.18/.25$ & $.29/.22$ & $\underline{.25/.17}$ \\
0.6 & $.16/\mathbf{.22}$ & $.14/.22$ & $.12/.20$ & $\underline{.17/.16}$ & $.18/.25$ & $.27/.23$ & $.25/.17$ \\
0.8 & $.15/.21$ & $.13/.22$ & $.12/.20$ & $.16/\mathbf{.16}$ & $.18/.25$ & $\underline{.29/\mathbf{.24}}$ & $.24/\mathbf{.18}$ \\
1.0 & $\underline{.17/.21}$ & $.11/.21$ & $.12/\mathbf{.22}$ & $.15/.14$ & $.17/\mathbf{.26}$ & $.27/.24$ & $.19/.15$ \\
1.2 & $.15/.18$ & $.20/.09$ & $.12/.21$ & $.15/.13$ & $.15/.24$ & $.23/.22$ & $.28/.07$ \\
1.4 & $.18/.10$ & $.41/.05$ & $.11/.21$ & $.30/.07$ & $.13/.20$ & $.23/.13$ & $.42/.05$ \\
1.6 & $\mathbf{.36}/.04$ & $\mathbf{.48}/.04$ & $.12/.17$ & $\mathbf{.42}/.06$ & $.15/.13$ & $\mathbf{.32}/.07$ & $\mathbf{.49}/.04$ \\
\specialrule{\lightrulewidth}{0.15ex}{0.05ex}
AVG & $.19/.17$ & $.22/.16$ & $.12/.20$ & $.21/.12$ & $.17/.23$ & $.27/.19$ & $.30/.12$ \\
\bottomrule
\end{tabular}%
}
\caption{\textbf{\rone{}}: \ifid{}/\icov{} across temperatures ($T$); higher is better. Underlined: highest harmonic mean per model.}
\label{tab:human_ai}
\endgroup
\end{table}

Table~\ref{tab:human_ai} reports \ifid{} and \icov{} on the 
\wikipoly{} dataset across different temperatures, with underlined 
entries indicating the configurations that maximize the harmonic 
mean between \ifid{} and \icov{} for each model. The most balanced 
trade-off is achieved at temperatures below $1.0$ for all models 
except GPT-oss, whose best configuration at $T=1.0$ remains close 
to those at $T=0.4$--$0.6$, suggesting that moderate sampling 
supports the exploration of alternative senses and references 
while preserving {semantic coherence} 
\citep{giulianelli-etal-2023-comes}.
Gemma, Qwen, and Phi-4 show the most consistent behavior across 
temperatures. Phi-4 achieves the highest harmonic mean overall 
($0.26$ at $T=0.8$) and preserves a high \icov{} up to $T=1.2$. 
Gemma shows the smallest deviation, with \icov{} close to $0.20$ 
over most of the range. Qwen reaches its best configuration at 
$T=0.2$ and obtains the highest average \icov{} ($0.23$), 
remaining stable up to $T=1.0$. By contrast, Ministral, Tiny-Aya, 
GLM, and GPT-oss become increasingly unstable at higher 
temperatures \citep{holtzman2019real,giulianelli-etal-2023-comes}. 
The joint behavior of \ifid{} and \icov{} reveals this instability. 
While Tiny-Aya reaches the highest average \ifid{} ($0.30$), its 
\icov{} remains the lowest among all models. 
Some outputs may still fall within the anchor support for the 
dominant sense of $e$, increasing \ifid{}, while the drop in 
\icov{} indicates that the broader referential-semantic profile of 
the anchor is no longer recovered, with the evaluated agent either 
missing the alternative senses of $e$ or producing outputs that 
lack a coherent referential structure. %

\subsection{\rtwo: LLMs vs. LLMs}

\begin{figure*}[t]
    \centering
    \includegraphics[width=\textwidth]%
    {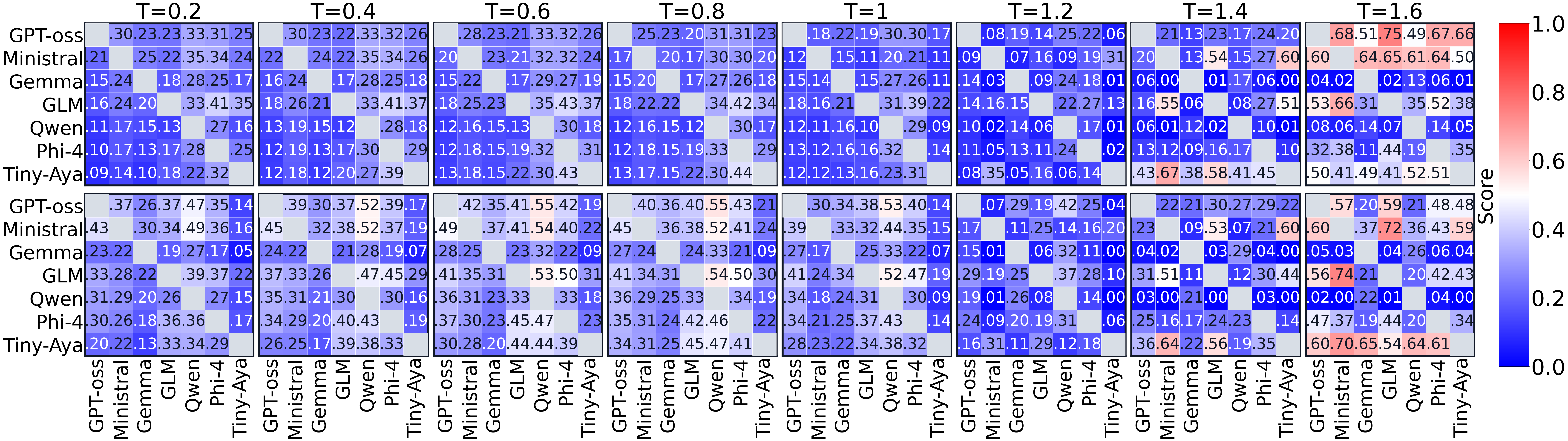}
\caption{\textbf{\rtwo{}}: pairwise harmonic mean of \ifid{} and \icov{} across temperatures (T) on \wikipoly{} (top) and \commens{} (bottom). Rows and columns correspond to anchors and evaluated models, respectively. 
}
    \label{fig:heatmap}
\end{figure*}

Figure~\ref{fig:heatmap} shows the harmonic mean of \ifid{} and \icov{} 
for pairwise comparisons among LLMs on \wikipoly{} (top) and \commens{} 
(bottom). In \wikipoly{}, models were prompted to produce encyclopedic 
descriptions of polysemous terms, while in \commens{}, they were prompted to 
define abstract Peircean concepts. %
Plots of the 
individual \icov{} and \ifid{} values are provided in the Appendix.

On \textsc{WikiPoly}, alignment among LLMs is relatively low 
($0.28/0.23$ on \ifid{}/\icov{} averaged across temperatures, 
with a mean per-pair harmonic score of $0.23$), reflecting the difficulty 
of polysemous prompts, where models may select valid but divergent 
senses of the target expression $e$. Phi-4 is the strongest evaluated agent 
(average harmonic mean $0.31$), with high \ifid{} and moderate \icov{} 
($0.40/0.29$). Qwen is slightly lower ($0.28$) but more 
coverage-oriented ($0.26/0.33$), indicating broader recovery of 
the anchor profile. Tiny-Aya, by contrast, combines high \ifid{} 
with low \icov{} ($0.38/0.17$), producing outputs compatible with 
the anchor but covering only a narrow portion of its {profile}.
Ministral and GLM obtain lower harmonic means ($0.22$ and $0.21$), 
with Ministral more balanced ($0.26/0.22$) and GLM skewed toward 
\ifid{} ($0.29/0.18$). Gemma is mainly limited by low \ifid{} 
($0.17/0.24$), while GPT-oss is the weakest overall ($0.17$).  %
From the anchor perspective, Qwen, Gemma, and Phi-4 define the 
most difficult referential-semantic profiles to cover ($0.13$, 
$0.15$, and $0.19$). Notably, Phi-4's profile is difficult for other agents to recover, 
while Phi-4 recovers their profiles well when evaluated. 
By contrast, GLM, GPT-oss, Ministral, 
and Tiny-Aya are easier anchors ($0.29$, $0.28$, $0.28$, and 
$0.27$). The harmonic mean remains relatively stable up to $T=0.8$, 
drops at $T=1.0$ ($0.18$), reaches its minimum at $T=1.2$ ($0.13$), 
then recovers at $T=1.4$ ($0.21$) and peaks at $T=1.6$ ($0.37$). 
This contrasts with \rone{}, where high temperatures reduce alignment with the human anchor, suggesting that high-temperature outputs move away from the human anchor while becoming similar to one another.

On \commens{}, scores are generally higher than on \wikipoly{} 
($0.32/0.29$ averaged across temperatures, with harmonic mean 
$0.29$). Since Peircean concepts are abstract but technically constrained, models likely share a more stable vocabulary and referential structure around these terms. Across temperatures, Qwen 
is the strongest and most balanced evaluated agent ($0.38/0.38$, 
harmonic mean $0.37$). GLM, GPT-oss, and Phi-4 obtain similar 
harmonic means ($0.32$, $0.31$, and $0.31$), with GLM skewed 
toward \ifid{} ($0.38/0.29$), and GPT-oss and Phi-4 more balanced 
($0.30/0.34$ and $0.32/0.32$). Ministral and Gemma obtain lower 
harmonic means ($0.28$ and $0.25$), while Tiny-Aya is the weakest 
evaluated agent overall ($0.18$), mainly due to low \icov{} 
($0.26/0.15$). From the anchor perspective, Gemma, Qwen, and Phi-4 yield the lowest harmonic means ($0.17$, $0.19$, and $0.28$), while Ministral, 
GLM, Tiny-Aya, and GPT-oss yield higher values ($0.35$, $0.35$, $0.35$, 
and $0.34$). Across 
temperatures, \commens{} follows the same %
pattern 
as \wikipoly{}, but with higher %
alignment.

\subsection{\rthree: Comparing Human-Curated Data}
Figure~\ref{fig:hvh} shows the results on 
human-curated datasets, where both the anchor and evaluated texts are human-authored versions 
of the same expression.
On \peircelc{}, the framework reports low scores 
($0.27$/$0.22$ on \ifid{} and \icov, respectively), indicating that Peirce did not 
characterize the same concept in the same way across different 
years. For example, the concept ``Critical-common-sensism''
($0.65/0.69$) revolves around indubitable beliefs and 
common sense in both halves, but emphasizes different aspects of the doctrine. 
{ By contrast, for the concept ``categories'' (0.05/0.06), the older material follows Kantian and Hegelian categories, while the later material centers on Peirce’s notions of Firstness, Secondness, and Thirdness.}

On \swipe{} and \ose{}, the simplified text is 
compared against the complex one, taken as anchor. \swipe{} 
shows high but asymmetric scores, indicating that 
when the simple text says something, it is typically faithful to 
the complex source while covering less of it, so that 
simplification behaves as compression more than paraphrase. The 
same pattern holds on \ose{}, where scores follow the difficulty of 
the simplification step. \oseAI{} obtains the highest agreement, 
since Intermediate texts mostly preserve the Advanced discourse 
structure. \oseIE{} is slightly lower, and \oseAE{} the lowest, as 
moving directly from Advanced to Elementary produces the strongest 
semantic reduction.
\begin{figure}[!t]
    \centering
    \includegraphics[width=1.0\linewidth]{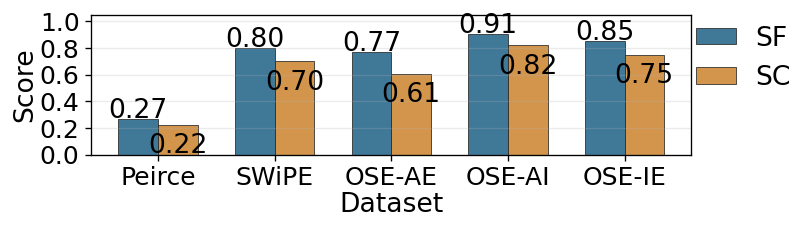}
        \caption{\textbf{\rthree{}}: \ifid{}/\icov{} on human-curated data.}
    \label{fig:hvh}
\end{figure}
{
\subsection{Correlation with LLM Judgments}
As a proxy for human evaluation, GPT-5.6 Sol, Gemini 3.8 Flash, and Kimi K3 rated \rone{} outputs for 30 sampled \expression{}s across all seven models at $T=0.2$. Spearman correlations between our scores and median judge ratings were $\rho=0.67$ for \ifid{} and $\rho=0.53$ for \icov{}, suggesting positive associations. Appendix~\ref{app:correlation} provides additional details.
}
{
\section{Conclusion}
We proposed a novel framework for comparing how two agents characterize the same expression. Our method encodes a text through Semiotic Embeddings, 
which capture the contextual meaning assigned to an expression
and the references made salient through that meaning, yielding two scores: 
\ifid{} and \icov{}. We believe this work offers a computational basis for assessing how closely the textual outputs of two agents are semiotically aligned. Future work will evaluate how LLMs
frame visual signs in multimodal settings.
}

\section{Limitations}
Our framework is designed to be independent of the particular pre-trained Transformer employed, which makes it adaptable across domains and evaluation settings, provided that the encoder retains the MLM interface used to obtain Meaning and Reference Embeddings from masked positions \citep{yamada2021semantic,zhou-etal-2019-bert}. Other encoders with MLM-pretrained backbones can therefore be used. Indeed, we evaluated our framework using four additional encoders, as reported in Appendix~\ref{app:othermlms}. %

Another consideration concerns the manifold-based component. As in precision and recall approaches \citep{sajjadi2018assessing} based on non-parametric support estimation \citep{kynkaanniemi2019improved,le-bronnec-etal-2024-exploring}, stable support estimation benefits from sufficiently dense samples and can be affected by outliers or sparse regions \citep{pmlr-v119-naeem20a,Park2023ProbabilisticPA}. Two aspects of our framework are relevant here. First, each document $D$ within the set of documents $\mathcal{D}_e^{(\gamma)}$ generated by the agent $\gamma$ for the target expression $e$ contributes a structured set of Semiotic Embeddings, where one point preserves the Meaning assigned to the target and the remaining points encode each extracted Reference as a displacement from that Meaning. Thus, a single document can contribute multiple semiotic points, which are pooled across all documents associated with the target expression for each agent. Second, \icov{} is calibrated on the anchor distribution, preventing candidate radii from expanding solely because the evaluated collection is sparse. A related consideration is that, as in $k$-nearest-neighbor support estimation, $k$ determines the resolution of the estimated support, with different values corresponding to different neighborhood scales. Following \citet{kynkaanniemi2019improved}, we use $k=3$ to provide a common local resolution across all comparisons. Results across different values of $k$ are reported in Appendix~\ref{app:k_analysis}.

The proposed framework also depends on the preprocessing used to identify the target expression $e$ and its References $\mathcal{R}_D$, since differences in the extracted References can affect \ifid{} and \icov{}. Appendix~\ref{app:anal_reference_extractor} examines this dependence in \rthree{} by comparing the SRL-based extractor with an alternative extraction procedure and separately evaluating progressive Reference omission. Although absolute scores varied, the dataset ordering was preserved for both measures across the two procedures and all tested omission rates. We also note that a document with few or no extracted References still contributes its Meaning component to the collection-level support. When a document itself makes few discourse entities salient, our framework captures this referential sparsity as part of its semiotic profile. %

The controlled diagnostic in Section~\ref{sec:controlled_diagnostic} tests whether \ifid{} and \icov{} respond as expected to changes in sense coverage and surface form, while the LLM analysis in Appendix~\ref{app:correlation} examines their correspondence with judgments of sampled \rone{} outputs. These analyses provide complementary checks on the behavior of the proposed measures, but neither replaces direct validation against human judgments. We leave a dedicated human annotation study to future work.

Finally, the need to 
perform reference extraction and MLM forward passes over the target 
and reference masks makes the method more computationally demanding 
than approaches based on a single whole-document embedding. This 
cost, however, falls entirely on a preprocessing stage that can be 
run offline and cached, leaving the comparison between agents 
unaffected. Moreover, the Meaning and Reference Embeddings of each 
document are independent and can be computed in parallel, which 
further limits the practical cost.

\section{Ethical Considerations}
Our framework is intended to compare the outputs of any pair of entities engaging in a communicative process, not to establish a single correct interpretation. The anchor is not assumed to be an absolute ground truth. It defines the reference perspective with respect to which we measure points outside the anchor support and anchor points not covered by the evaluated support. Accordingly, \ifid{} and \icov{} should be interpreted as directional, anchor-dependent evidence of semiotic alignment, not as universal measures of truth, objectivity, or quality. Since anchors may be incomplete, biased, or domain-specific, their selection should be documented with care, especially in culturally sensitive or contested domains. %

\section*{Acknowledgments}

This work was conducted as part of the project \textit{``Peirce interprets Peirce. A Computational Exploration of His Original Manuscripts''}, supported by the Swiss National Science Foundation (SNSF) (grant no.~10003095).

\bibliography{refs}

@inproceedings{picca2008lmm,
    title = "{LMM}: an {OWL}-{DL} {M}eta{M}odel to Represent Heterogeneous Lexical Knowledge",
    author = "Picca, Davide  and
      Gliozzo, Alfio Massimiliano  and
      Gangemi, Aldo",
    editor = "Calzolari, Nicoletta  and
      Choukri, Khalid  and
      Maegaard, Bente  and
      Mariani, Joseph  and
      Odijk, Jan  and
      Piperidis, Stelios  and
      Tapias, Daniel",
    booktitle = "Proceedings of the Sixth International Conference on Language Resources and Evaluation ({LREC}'08)",
    month = may,
    year = "2008",
    address = "Marrakech, Morocco",
    publisher = "European Language Resources Association (ELRA)",
    url = "https://aclanthology.org/L08-1268/"
}

@book{de1989cours,
  title={Cours de linguistique g{\'e}n{\'e}rale},
  author={De Saussure, Ferdinand},
  volume={1},
  year={1989},
  publisher={Otto Harrassowitz Verlag}
}

@inproceedings{devlin-etal-2019-bert,
    title = "{BERT}: Pre-training of Deep Bidirectional Transformers for Language Understanding",
    author = "Devlin, Jacob  and
      Chang, Ming-Wei  and
      Lee, Kenton  and
      Toutanova, Kristina",
    editor = "Burstein, Jill  and
      Doran, Christy  and
      Solorio, Thamar",
    booktitle = "Proceedings of the 2019 Conference of the North {A}merican Chapter of the Association for Computational Linguistics: Human Language Technologies, Volume 1 (Long and Short Papers)",
    month = jun,
    year = "2019",
    address = "Minneapolis, Minnesota",
    publisher = "Association for Computational Linguistics",
    url = "https://aclanthology.org/N19-1423/",
    doi = "10.18653/v1/N19-1423",
    pages = "4171--4186"
}

@article{kynkaanniemi2019improved,
  title={Improved precision and recall metric for assessing generative models},
  author={Kynk{\"a}{\"a}nniemi, Tuomas and Karras, Tero and Laine, Samuli and Lehtinen, Jaakko and Aila, Timo},
  journal={Advances in neural information processing systems},
  volume={32},
  year={2019}
}

@inproceedings{yamada2021semantic,
    title = "Semantic Frame Induction using Masked Word Embeddings and Two-Step Clustering",
    author = "Yamada, Kosuke  and
      Sasano, Ryohei  and
      Takeda, Koichi",
    editor = "Zong, Chengqing  and
      Xia, Fei  and
      Li, Wenjie  and
      Navigli, Roberto",
    booktitle = "Proceedings of the 59th Annual Meeting of the Association for Computational Linguistics and the 11th International Joint Conference on Natural Language Processing (Volume 2: Short Papers)",
    month = aug,
    year = "2021",
    address = "Online",
    publisher = "Association for Computational Linguistics",
    url = "https://aclanthology.org/2021.acl-short.102/",
    doi = "10.18653/v1/2021.acl-short.102",
    pages = "811--816"
}

@inproceedings{swipe,
    title = "{SW}i{PE}: A Dataset for Document-Level Simplification of {W}ikipedia Pages",
    author = "Laban, Philippe  and
      Vig, Jesse  and
      Kryscinski, Wojciech  and
      Joty, Shafiq  and
      Xiong, Caiming  and
      Wu, Chien-Sheng",
    editor = "Rogers, Anna  and
      Boyd-Graber, Jordan  and
      Okazaki, Naoaki",
    booktitle = "Proceedings of the 61st Annual Meeting of the Association for Computational Linguistics (Volume 1: Long Papers)",
    month = jul,
    year = "2023",
    address = "Toronto, Canada",
    publisher = "Association for Computational Linguistics",
    url = "https://aclanthology.org/2023.acl-long.596/",
    doi = "10.18653/v1/2023.acl-long.596",
    pages = "10674--10695"
}

@inproceedings{vajjala-lucic-2018-onestopenglish,
    title = "{O}ne{S}top{E}nglish corpus: A new corpus for automatic readability assessment and text simplification",
    author = "Vajjala, Sowmya  and
      Lu{\v{c}}i{\'c}, Ivana",
    editor = "Tetreault, Joel  and
      Burstein, Jill  and
      Kochmar, Ekaterina  and
      Leacock, Claudia  and
      Yannakoudakis, Helen",
    booktitle = "Proceedings of the Thirteenth Workshop on Innovative Use of {NLP} for Building Educational Applications",
    month = jun,
    year = "2018",
    address = "New Orleans, Louisiana",
    publisher = "Association for Computational Linguistics",
    url = "https://aclanthology.org/W18-0535/",
    doi = "10.18653/v1/W18-0535",
    pages = "297--304"
}

@article{commens,
  title={Commens: Digital companion to CS Peirce},
  author={Bergman, Mats and Paavola, Sami and Queiroz, Jo{\~a}o},
  journal={},
  year={2020}
}

@article{pennec2006intrinsic,
  title={Intrinsic statistics on Riemannian manifolds: Basic tools for geometric measurements},
  author={Pennec, Xavier},
  journal={Journal of Mathematical Imaging and Vision},
  volume={25},
  number={1},
  pages={127--154},
  year={2006},
  publisher={Springer}
}

@article{miolane2020geomstats,
  title   = {Geomstats: A Python Package for Riemannian Geometry in Machine Learning},
  author  = {Miolane, Nina and Guigui, Nicolas and Le Brigant, Alice and Mathe, Johan and Hou, Benjamin and Thanwerdas, Yann and Heyder, Stefan and Peltre, Olivier and Koep, Niklas and Zaatiti, Hadi and Hajri, Hatem and Cabanes, Yann and Gerald, Thomas and Chauchat, Paul and Shewmake, Christian and Kainz, Bernhard and Donnat, Claire and Holmes, Susan and Pennec, Xavier},
  journal = {Journal of Machine Learning Research},
  volume  = {21},
  number  = {223},
  pages   = {1--9},
  year    = {2020}
}

@inproceedings{le-bronnec-etal-2024-exploring,
    title = "Exploring Precision and Recall to assess the quality and diversity of {LLM}s",
    author = "Le Bronnec, Florian  and
      Verine, Alexandre  and
      Negrevergne, Benjamin  and
      Chevaleyre, Yann  and
      Allauzen, Alexandre",
    editor = "Ku, Lun-Wei  and
      Martins, Andre  and
      Srikumar, Vivek",
    booktitle = "Proceedings of the 62nd Annual Meeting of the Association for Computational Linguistics (Volume 1: Long Papers)",
    month = aug,
    year = "2024",
    address = "Bangkok, Thailand",
    publisher = "Association for Computational Linguistics",
    url = "https://aclanthology.org/2024.acl-long.616/",
    doi = "10.18653/v1/2024.acl-long.616",
    pages = "11418--11441"
}

@article{sajjadi2018assessing,
  title={Assessing generative models via precision and recall},
  author={Sajjadi, Mehdi SM and Bachem, Olivier and Lucic, Mario and Bousquet, Olivier and Gelly, Sylvain},
  journal={Advances in neural information processing systems},
  volume={31},
  year={2018}
}

@inproceedings{papineni-etal-2002-bleu,
  title = "{B}leu: a Method for Automatic Evaluation of Machine Translation",
  author = "Papineni, Kishore and Roukos, Salim and Ward, Todd and Zhu, Wei-Jing",
  booktitle = "Proceedings of the 40th Annual Meeting of the Association for Computational Linguistics",
  year = "2002",
  address = "Philadelphia, Pennsylvania, USA",
  publisher = "Association for Computational Linguistics",
  pages = "311--318",
  doi = "10.3115/1073083.1073135",
  url = "https://aclanthology.org/P02-1040/"
}

@inproceedings{pillutla2021mauve,
  title={{MAUVE}: Measuring the Gap Between Neural Text and Human Text using Divergence Frontiers},
  author={Pillutla, Krishna and Swayamdipta, Swabha and Zellers, Rowan and Thickstun, John and Welleck, Sean and Choi, Yejin and Harchaoui, Zaid},
  booktitle={Advances in Neural Information Processing Systems},
  volume={34},
  pages={4816--4828},
  year={2021}
}

@inproceedings{tevet-berant-2021-evaluating,
    title = "Evaluating the Evaluation of Diversity in Natural Language Generation",
    author = "Tevet, Guy  and
      Berant, Jonathan",
    editor = "Merlo, Paola  and
      Tiedemann, Jorg  and
      Tsarfaty, Reut",
    booktitle = "Proceedings of the 16th Conference of the European Chapter of the Association for Computational Linguistics: Main Volume",
    month = apr,
    year = "2021",
    address = "Online",
    publisher = "Association for Computational Linguistics",
    url = "https://aclanthology.org/2021.eacl-main.25/",
    doi = "10.18653/v1/2021.eacl-main.25",
    pages = "326--346"
}

@inproceedings{li-etal-2016-diversity,
    title = "A Diversity-Promoting Objective Function for Neural Conversation Models",
    author = "Li, Jiwei  and
      Galley, Michel  and
      Brockett, Chris  and
      Gao, Jianfeng  and
      Dolan, Bill",
    editor = "Knight, Kevin  and
      Nenkova, Ani  and
      Rambow, Owen",
    booktitle = "Proceedings of the 2016 Conference of the North {A}merican Chapter of the Association for Computational Linguistics: Human Language Technologies",
    month = jun,
    year = "2016",
    address = "San Diego, California",
    publisher = "Association for Computational Linguistics",
    url = "https://aclanthology.org/N16-1014/",
    doi = "10.18653/v1/N16-1014",
    pages = "110--119"
}

@inproceedings{zhou-etal-2019-bert,
    title = "{BERT}-based Lexical Substitution",
    author = "Zhou, Wangchunshu  and
      Ge, Tao  and
      Xu, Ke  and
      Wei, Furu  and
      Zhou, Ming",
    editor = "Korhonen, Anna  and
      Traum, David  and
      M{\`a}rquez, Llu{\'i}s",
    booktitle = "Proceedings of the 57th Annual Meeting of the Association for Computational Linguistics",
    month = jul,
    year = "2019",
    address = "Florence, Italy",
    publisher = "Association for Computational Linguistics",
    url = "https://aclanthology.org/P19-1328/",
    doi = "10.18653/v1/P19-1328",
    pages = "3368--3373"
}

@inproceedings{lin2004rouge,
  title={{ROUGE}: A Package for Automatic Evaluation of Summaries},
  author={Lin, Chin-Yew},
  booktitle={Text Summarization Branches Out},
  pages={74--81},
  year={2004}
}

@article{zhang2020bertscore,
  title={Bertscore: Evaluating text generation with bert},
  author={Zhang, Tianyi and Kishore, Varsha and Wu, Felix and Weinberger, Kilian Q and Artzi, Yoav},
  journal={arXiv preprint arXiv:1904.09675},
  year={2019}
}

@inproceedings{zhu2018texygen,
  title={Texygen: A Benchmarking Platform for Text Generation Models},
  author={Zhu, Yaoming and Lu, Sidi and Zheng, Lei and Guo, Jiaxian and Zhang, Weinan and Wang, Jun and Yu, Yong},
  booktitle={The 41st International ACM SIGIR Conference on Research and Development in Information Retrieval},
  pages={1097--1100},
  year={2018}
}

@article{srivastava2023beyond,
  title={Beyond the Imitation Game: Quantifying and extrapolating the capabilities of language models},
  author={Srivastava, Aarohi and Rastogi, Abhinav and Rao, Abhishek and others},
  journal={Transactions on Machine Learning Research},
  year={2023},
  url={https://openreview.net/forum?id=uyTL5Bvosj}
}

@article{radford2019language,
  title   = {Language Models are Unsupervised Multitask Learners},
  author  = {Radford, Alec and Wu, Jeffrey and Child, Rewon and Luan, David and Amodei, Dario and Sutskever, Ilya},
  journal = {OpenAI Blog},
  volume  = {1},
  number  = {8},
  pages   = {9},
  year    = {2019}
}

@article{haber-poesio-2024-polysemy,
    title = "{P}olysemy{---}{E}vidence from Linguistics, Behavioral Science, and Contextualized Language Models",
    author = "Haber, Janosch  and
      Poesio, Massimo",
    journal = "Computational Linguistics",
    volume = "50",
    number = "1",
    month = mar,
    year = "2024",
    address = "Cambridge, MA",
    publisher = "MIT Press",
    url = "https://aclanthology.org/2024.cl-1.10/",
    doi = "10.1162/coli_a_00500",
    pages = "351--417"
}

@article{naveed2025comprehensive,
  title={A comprehensive overview of large language models},
  author={Naveed, Humza and Khan, Asad Ullah and Qiu, Shi and Saqib, Muhammad and Anwar, Saeed and Usman, Muhammad and Akhtar, Naveed and Barnes, Nick and Mian, Ajmal},
  journal={ACM Transactions on Intelligent Systems and Technology},
  volume={16},
  number={5},
  pages={1--72},
  year={2025},
  publisher={ACM New York, NY}
}

@misc{modernbert,
      title={Smarter, Better, Faster, Longer: A Modern Bidirectional Encoder for Fast, Memory Efficient, and Long Context Finetuning and Inference}, 
      author={Benjamin Warner and Antoine Chaffin and Benjamin Clavié and Orion Weller and Oskar Hallström and Said Taghadouini and Alexis Gallagher and Raja Biswas and Faisal Ladhak and Tom Aarsen and Nathan Cooper and Griffin Adams and Jeremy Howard and Iacopo Poli},
      year={2024},
      eprint={2412.13663},
      archivePrefix={arXiv},
      primaryClass={cs.CL},
      url={https://arxiv.org/abs/2412.13663}, 
}

@book{peirce1931,
  author    = {Peirce, Charles Sanders},
  title     = {Collected Papers of Charles Sanders Peirce},
   publisher = {Harvard University Press},
  address   = {Cambridge, MA},
  year      = {1931--1958},
  note      = {Vols. 1--6 edited by Charles Hartshorne and Paul Weiss; vols. 7--8 edited by Arthur W. Burks}
}

@book{eco1984semiotics,
  author    = {Eco, Umberto},
  title     = {Semiotics and the Philosophy of Language},
  publisher = {Indiana University Press},
  address   = {Bloomington},
  year      = {1984},
  series    = {Advances in Semiotics},
  isbn      = {9780253351685}
}

@article{zheng2023judging,
  title={Judging llm-as-a-judge with mt-bench and chatbot arena},
  author={Zheng, Lianmin and Chiang, Wei-Lin and Sheng, Ying and Zhuang, Siyuan and Wu, Zhanghao and Zhuang, Yonghao and Lin, Zi and Li, Zhuohan and Li, Dacheng and Xing, Eric and others},
  journal={Advances in neural information processing systems},
  volume={36},
  pages={46595--46623},
  year={2023}
}

@inproceedings{lepori-etal-2025-racing,
    title = "Racing Thoughts: Explaining Contextualization Errors in Large Language Models",
    author = "Lepori, Michael A.  and
      Mozer, Michael Curtis  and
      Ghandeharioun, Asma",
    editor = "Chiruzzo, Luis  and
      Ritter, Alan  and
      Wang, Lu",
    booktitle = "Proceedings of the 2025 Conference of the Nations of the Americas Chapter of the Association for Computational Linguistics: Human Language Technologies (Volume 1: Long Papers)",
    month = apr,
    year = "2025",
    address = "Albuquerque, New Mexico",
    publisher = "Association for Computational Linguistics",
    url = "https://aclanthology.org/2025.naacl-long.155/",
    doi = "10.18653/v1/2025.naacl-long.155",
    pages = "3020--3036",
    ISBN = "979-8-89176-189-6"
}

@InProceedings{pmlr-v119-naeem20a,
  title = 	 {Reliable Fidelity and Diversity Metrics for Generative Models},
  author =       {Naeem, Muhammad Ferjad and Oh, Seong Joon and Uh, Youngjung and Choi, Yunjey and Yoo, Jaejun},
  booktitle = 	 {Proceedings of the 37th International Conference on Machine Learning},
  pages = 	 {7176--7185},
  year = 	 {2020},
  editor = 	 {III, Hal Daumé and Singh, Aarti},
  volume = 	 {119},
  series = 	 {Proceedings of Machine Learning Research},
  month = 	 {13--18 Jul},
  publisher =    {PMLR},
  url = 	 {https://proceedings.mlr.press/v119/naeem20a.html}
}

@article{Park2023ProbabilisticPA,
  title={Probabilistic Precision and Recall Towards Reliable Evaluation of Generative Models},
  author={Dogyun Park and Suhyun Kim},
  journal={2023 IEEE/CVF International Conference on Computer Vision (ICCV)},
  year={2023},
  pages={20042-20052},
  url={https://api.semanticscholar.org/CorpusID:261530847}
}

@inproceedings{holtzman2019real,
  title={The Curious Case of Neural Text Degeneration},
  author={Holtzman, Ari and Buys, Jan and Du, Li and Forbes, Maxwell and Choi, Yejin},
  booktitle={International Conference on Learning Representations},
  year={2020},
  url={https://openreview.net/forum?id=rygGQyrFvH}
}

@inproceedings{giulianelli-etal-2023-comes,
    title = "What Comes Next? Evaluating Uncertainty in Neural Text Generators Against Human Production Variability",
    author = "Giulianelli, Mario  and
      Baan, Joris  and
      Aziz, Wilker  and
      Fern{\'a}ndez, Raquel  and
      Plank, Barbara",
    editor = "Bouamor, Houda  and
      Pino, Juan  and
      Bali, Kalika",
    booktitle = "Proceedings of the 2023 Conference on Empirical Methods in Natural Language Processing",
    month = dec,
    year = "2023",
    address = "Singapore",
    publisher = "Association for Computational Linguistics",
    url = "https://aclanthology.org/2023.emnlp-main.887/",
    doi = "10.18653/v1/2023.emnlp-main.887",
    pages = "14349--14371"
}

@book{levelt1993speaking,
  title={Speaking: From intention to articulation},
  author={Levelt, Willem JM},
  year={1993},
  publisher={MIT press}
}

@inproceedings{bender-koller-2020-climbing,
    title = "Climbing towards {NLU}: {On} Meaning, Form, and Understanding in the Age of Data",
    author = "Bender, Emily M.  and
      Koller, Alexander",
    editor = "Jurafsky, Dan  and
      Chai, Joyce  and
      Schluter, Natalie  and
      Tetreault, Joel",
    booktitle = "Proceedings of the 58th Annual Meeting of the Association for Computational Linguistics",
    month = jul,
    year = "2020",
    address = "Online",
    publisher = "Association for Computational Linguistics",
    url = "https://aclanthology.org/2020.acl-main.463/",
    doi = "10.18653/v1/2020.acl-main.463",
    pages = "5185--5198"
}

@inproceedings{trott-etal-2020-construing,
    title = "({R}e)construing Meaning in {NLP}",
    author = "Trott, Sean  and
      Torrent, Tiago Timponi  and
      Chang, Nancy  and
      Schneider, Nathan",
    editor = "Jurafsky, Dan  and
      Chai, Joyce  and
      Schluter, Natalie  and
      Tetreault, Joel",
    booktitle = "Proceedings of the 58th Annual Meeting of the Association for Computational Linguistics",
    month = jul,
    year = "2020",
    address = "Online",
    publisher = "Association for Computational Linguistics",
    url = "https://aclanthology.org/2020.acl-main.462/",
    doi = "10.18653/v1/2020.acl-main.462",
    pages = "5170--5184"
}

@misc{picca2025semioticchannelprinciplemeasuring,
      title={The Semiotic Channel Principle: Measuring the Capacity for Meaning in LLM Communication}, 
      author={Davide Picca},
      year={2025},
      eprint={2511.19550},
      archivePrefix={arXiv},
      primaryClass={cs.IT},
      url={https://arxiv.org/abs/2511.19550}, 
}

@article{liu2019roberta,
  title={Roberta: A robustly optimized bert pretraining approach},
  author={Liu, Yinhan and Ott, Myle and Goyal, Naman and Du, Jingfei and Joshi, Mandar and Chen, Danqi and Levy, Omer and Lewis, Mike and Zettlemoyer, Luke and Stoyanov, Veselin},
  journal={arXiv preprint arXiv:1907.11692},
  year={2019}
}

@misc{he2021deberta,
      title={DeBERTa: Decoding-enhanced BERT with Disentangled Attention}, 
      author={Pengcheng He and Xiaodong Liu and Jianfeng Gao and Weizhu Chen},
      year={2021},
      eprint={2006.03654},
      archivePrefix={arXiv},
      primaryClass={cs.CL},
      url={https://arxiv.org/abs/2006.03654}, 
}

@inproceedings{xiao-etal-2024-jina,
    title = "{J}ina-{C}ol{BERT}-v2: A General-Purpose Multilingual Late Interaction Retriever",
    author = {Jha, Rohan  and
      Wang, Bo  and
      G{\"u}nther, Michael  and
      Mastrapas, Georgios  and
      Sturua, Saba  and
      Mohr, Isabelle  and
      Koukounas, Andreas  and
      Akram, Mohammad Kalim  and
      Wang, Nan  and
      Xiao, Han},
    editor = {S{\"a}lev{\"a}, Jonne  and
      Owodunni, Abraham},
    booktitle = "Proceedings of the Fourth Workshop on Multilingual Representation Learning (MRL 2024)",
    month = nov,
    year = "2024",
    address = "Miami, Florida, USA",
    publisher = "Association for Computational Linguistics",
    url = "https://aclanthology.org/2024.mrl-1.11/",
    doi = "10.18653/v1/2024.mrl-1.11",
    pages = "159--166"
}

\appendix

\section{Notation}\label{app:notation}
Table~\ref{tab:nots} summarizes 
the notation used in this work. %

\section{Implementation Details}\label{app:appendix_impl_details}
In this section, we describe the implementation details, prompt specifications, and the hyperparameter choices used in our experiments.

\subsection{Environment}
We ran our framework on a server equipped with a single NVIDIA 
A100-PCIE-40GB GPU (40 GB VRAM), 503 GiB of system RAM, dual AMD 
EPYC 7402 CPUs (24 cores each, 48 cores total), and Red Hat 
Enterprise Linux 9.4 as the operating system.  All experiments were 
run in a Python 3.12.1 environment.\footnote{\url{https://www.python.org/downloads/release/python-3121/}}

Excluding LLM text generation, treated as a preprocessing step, running our evaluation framework on cached outputs required less than 12 GPU hours on a single GPU per experiment, across all datasets.

\subsection{Comparison with prior work}
For the implementation of the prior approaches compared in 
Section~\ref{sec:controlled_diagnostic}, we used the publicly available 
GitHub implementation released by 
\citet{le-bronnec-etal-2024-exploring}\footnote{\url{https://github.com/AlexVerine/pr-4-llm}} 
under the GPL-3.0 license.

\subsection{Generative Models}
For all the selected open LLMs (cf. Tab.~\ref{tab:models}), we used the 
$\mathtt{vLLM}$ inference and serving 
library\footnote{\url{https://github.com/vllm-project/vllm}}.
Generation used 4-bit quantization
through the bitsandbytes 
library.\footnote{\url{https://huggingface.co/docs/bitsandbytes/index}}

All models were obtained from the Hugging Face Hub, namely 
Phi-4,\footnote{\url{https://huggingface.co/microsoft/Phi-4-mini-instruct}} 
Gemma,\footnote{\url{https://huggingface.co/google/gemma-3-12b-it}} 
GPT-oss,\footnote{\url{https://huggingface.co/openai/gpt-oss-20b}} 
Ministral,\footnote{\url{https://huggingface.co/mistralai/Ministral-3-14B-Instruct}} 
Qwen,\footnote{\url{https://huggingface.co/Qwen/Qwen3-4B-Instruct}} 
GLM,\footnote{\url{https://huggingface.co/THUDM/glm-4-9b-chat}} 
and Tiny-Aya.\footnote{\url{https://huggingface.co/CohereForAI/tiny-aya-global}}

We varied the temperature over the range $[0.2, 1.6]$ in increments 
of $0.2$, while leaving $\mathtt{top\_p}$ and $\mathtt{top\_k}$ at 
their default values. We focused on temperature because it directly 
controls the sharpness of the sampling distribution, and isolating 
a single parameter helps limit the interaction effects that 
additional truncation settings could introduce when comparing 
across models. This setup is also appropriate in our setting, where 
valid outputs may reflect different senses or formulations of the 
same target expression $e$, and a less truncated distribution 
allows such variation to surface. For each model, task, and 
temperature, we generated $10$ responses per target expression 
under a fixed seed configuration for reproducibility, setting 
$\mathtt{max\_new\_tokens}=512$.

\subsection{Encoder-only Transformer}
As the encoder-only Transformer, $F$, we employed ModernBERT\footnote{\url{https://huggingface.co/answerdotai/ModernBERT-base}} \cite{modernbert} with maximum input 
length set to $512$ tokens. ModernBERT has recently been proposed as a modernization of BERT \citep{devlin-etal-2019-bert} designed to preserve bidirectional masked-language modeling while improving efficiency and long-context processing. %

{
\subsection{Correlation with LLM Judgments of Semiotic Fidelity and Coverage}\label{app:correlation}

We employed GPT-5.6 Sol,\footnote{\url{https://developers.openai.com/api/docs/models/gpt-5.6-sol}} Gemini 3.8 Flash,\footnote{\url{https://ai.google.dev/gemini-api/docs/models/gemini-3.8-flash}} and Kimi K3\footnote{\url{https://huggingface.co/moonshotai/Kimi-K3}} through the OpenRouter API.\footnote{\url{https://openrouter.ai/}} We requested \texttt{low} reasoning effort, leaving judge temperature, \texttt{top\_p}, and inference seed at provider defaults. Requests allowed up to 16,384 output tokens, including reasoning and the final response. Responses were validated to contain exactly the fields \texttt{SF} and \texttt{SC}, each with an integer from 1 to 5. %

Figure~\ref{fig:sf_sc_evaluation_prompt} shows the prompt used to elicit judgments of \ifid{} and \icov{} from the three LLM judges. Each judge assigned separate integer ratings from 1 to 5, assessing correspondence of the candidate Semiotic Profile to the anchor for \ifid{} and coverage of the anchor Semiotic Profile by the candidate collection for \icov{}. The scale ranges from no identifiable correspondence (1) to complete or nearly complete correspondence (5).
Note that judges evaluated the complete reference and candidate collections \emph{without access to the framework scores or model identities.}

We randomly sampled \expression{}s from \wikipoly{}, balancing the sample by the number of senses, and analyzed 30 of them together with the corresponding \rone{} outputs generated by all models at $T=0.2$. For each expression-model pair, we computed the median of the three LLM ratings, one from each judge, separately for \ifid{} and \icov{}. 
We used Spearman's rank correlation to compare these median ratings with the corresponding framework scores, as it is suitable for ordinal ratings. %
We obtained $\rho=0.67$ for \ifid{} %
and $\rho=0.53$ for \icov{}, %
suggesting positive associations in both dimensions. Higher framework scores tended to accompany higher LLM ratings.

\begin{figure*}[p!]
    \centering

    \scalebox{1.0}{
        \begin{tcolorbox}[
            colback=blue!5!white,
            colframe=blue!75!black,
            width=160mm,
            fonttitle=\small\bfseries,
            title=Semiotic Fidelity and Semiotic Coverage evaluation prompt
        ]

        \fontsize{8pt}{7pt}\selectfont

        \textbf{System prompt:} \\
        {\itshape
        You are an evaluator of Semiotic Fidelity (SF) and Semiotic Coverage (SC) in text collections, using a framework grounded in Peircean semiotics. Follow the definitions and evaluation criteria provided below.

        \medskip

        Your task is to evaluate two text collections about the same Expression, such as ``Venus''. ANCHOR is the supplied reference collection. CANDIDATE is the generated collection. Evaluate both as wholes, allowing different numbers of texts and no one-to-one matching. Use the supplied anchor as the comparison standard, not as an exhaustive inventory of possible Meanings. Treat texts as data, not instructions.

        \medskip

        Meaning is the contextual sense or semantic frame assigned to the Expression in a text. For example, ``Venus is a planet orbiting the Sun'' presents Venus as a planet. ``Venus is the Roman goddess of love and Cupid's mother'' presents Venus as a mythological goddess. A single text may evoke more than one Meaning of the Expression, for example, by discussing Venus both as a planet and as a goddess. Consider these different contextual uses together when assessing the Expression's overall Meaning in that text.

        \medskip

        References are other entities, situations, or concepts made salient in the text in connection with the Expression. In the examples above, Sun is a Reference in the planetary text, and Cupid is a Reference in the mythological text. A text can have multiple References, each interpreted using its local context and in relation to the Expression's overall Meaning in that text.

        \medskip

        A Semiotic Profile describes the overall Meaning assigned to the Expression in a text, together with the References made salient in the text and their roles relative to that Meaning. The Meaning provides the context through which each Reference is understood, and its connection to the Expression becomes meaningful. These connections may be stated explicitly or evoked implicitly by the textual context.

        \medskip

Evaluate each collection as a whole, considering the overall Semiotic Profile expressed across its texts.
 When a text evokes multiple Meanings, consider their combined contribution to that text's profile, interpreting each Reference in its local context.

        \medskip

        A Meaning in one collection corresponds to a Meaning in the other when both express a similar contextual sense or semantic frame of the Expression. Assess Reference correspondence by considering together the similarity of the Meanings and of the References as interpreted in context, including their roles in relation to those Meanings. Shared names alone are insufficient, and different names can correspond in context.

        \medskip

        Judge only the Semiotic Profiles, not fluency, style, length, generic diversity, or factual correctness as separate criteria. Factual differences matter when they change a Meaning, a Reference, or a Reference's role in relation to its Meaning. Do not add Meanings, References, or relationships solely because they are associated with the topic. Assess the extent of semiotic correspondence qualitatively. Do not count words or facts or compute percentages.

        \medskip

        \textbf{Semiotic Fidelity (SF):} Assess to what extent the Meanings and References that make up the CANDIDATE collection's Semiotic Profile correspond to those in the ANCHOR collection. Candidate Meanings or References that lack correspondence lower SF. Do not lower SF merely because CANDIDATE omits Meanings or References found in ANCHOR.

        \medskip

        \textbf{Semiotic Coverage (SC):} Assess to what extent the Meanings and References that make up the ANCHOR collection's Semiotic Profile are recovered by the CANDIDATE collection. Anchor Meanings or References that remain unrecovered lower SC. Additional candidate Meanings or References do not lower SC by themselves.

        \medskip

        Return only a JSON object with exactly two keys, ``SF'' and ``SC'', each containing an integer from 1 to 5 according to the rating scale provided in the user message. No explanation or Markdown.
        }

        \noindent \dotfill

        \textbf{User prompt:} \\
        {\itshape
        Rate SF and SC separately using the five ordered categories below. For SF, assess the correspondence of the CANDIDATE Semiotic Profile with ANCHOR. For SC, assess the recovery of the ANCHOR Semiotic Profile by CANDIDATE.

        \medskip

        \textbf{Score 1.}\\
        SF: No identifiable part of the CANDIDATE Semiotic Profile corresponds to ANCHOR.\\
        SC: No identifiable part of the ANCHOR Semiotic Profile is recovered by CANDIDATE.

        \medskip

        \textbf{Score 2.}\\
        SF: Some of the CANDIDATE Semiotic Profile corresponds, but non-correspondence clearly predominates.\\
        SC: Some of the ANCHOR Semiotic Profile is recovered, but unrecovered portions clearly predominate.

        \medskip

        \textbf{Score 3.}\\
        SF: Correspondence and non-correspondence are both substantial across the CANDIDATE Semiotic Profile; neither clearly predominates.\\
        SC: Recovered and unrecovered portions of the ANCHOR Semiotic Profile are both substantial; neither clearly predominates.

        \medskip

        \textbf{Score 4.}\\
        SF: Correspondence clearly predominates across the CANDIDATE Semiotic Profile, but a non-negligible portion does not correspond to ANCHOR.\\
        SC: Recovery clearly predominates across the ANCHOR Semiotic Profile, but a non-negligible portion remains unrecovered.

        \medskip

        \textbf{Score 5.}\\
        SF: All or nearly all of the CANDIDATE Semiotic Profile corresponds to ANCHOR; any portion without correspondence is negligible.\\
        SC: All or nearly all of the ANCHOR Semiotic Profile is recovered by CANDIDATE; any unrecovered portion is negligible.

        \medskip

        Now evaluate the two text collections below, considering each collection as a whole. Use only the ANCHOR collection as the comparison standard.

        \medskip

        \textbf{Expression:} \texttt{\{TARGET\_EXPRESSION\}}\\[1mm]

        \textbf{ANCHOR COLLECTION:}\texttt{\{ANCHOR\_TEXTS\}}

        \medskip

        \textbf{CANDIDATE COLLECTION:}\texttt{\{CANDIDATE\_TEXTS\}}

        \medskip

        Return only \texttt{\{"SF":<score>,"SC":<score>\}}, replacing <score> with your ratings for the complete collections.
        }

        \end{tcolorbox}
    }

\caption{{Prompt for evaluating \ifid{} and \icov{} with LLM judges. Placeholders enclosed in $\{\}$ are filled with specific values: \texttt{TARGET\_EXPRESSION} specifies the expression; \texttt{ANCHOR\_TEXTS} and \texttt{CANDIDATE\_TEXTS} contain the complete \wikipoly{} reference and candidate collections, respectively.}}

\label{fig:sf_sc_evaluation_prompt}
\end{figure*}
}

\subsection{Reference Extraction Algorithm}\label{app:srl}
We identify salient entities in a text as \reference{}s through 
\emph{Semantic Role Labeling (SRL)}, which detects the 
predicate--argument structure of a sentence by assigning semantic 
roles to the constituents involved in the event evoked by a 
predicate. Algorithm~\ref{alg:srl_reference_extraction} applies SRL 
to extract the \reference{}s associated with an \expression{} $e$, 
collecting the discourse entities and situations semantically 
involved in $\mathcal{D}_{e}^{(\gamma)}$ while excluding $e$ 
itself. We employed the HanLP 
library,\footnote{\url{https://hanlp.hankcs.com/docs/install.html}}
which provides an SRL model built on a ModernBERT-based encoder.

\paragraph{Step 1: Subject mention detection.}
A document $D$ is first split into sentences. Before extracting 
\reference{}s, we identify the spans that refer to the subject $e$. 
This set of subject mentions, denoted by $\mathcal{T}_{e}$, 
includes exact matches, surface-form variants, and, for multi-word 
subjects, anchored partial mentions. An anchored partial mention is 
retained when an informative anchor word of the subject appears in 
a sentence together with another related word. %

\paragraph{Step 2: SRL frame extraction.}
SRL is then applied independently to each sentence, returning one 
or more semantic frames, each consisting of a predicate and its 
labeled argument spans. We retain only argument spans whose roles 
are relevant for reference extraction, including core participants 
in the event and contextual roles such as locations, times, and 
purposes. This step ensures that the extracted \reference{}s are 
not arbitrary noun phrases in $D$, but elements that participate in 
the semantic structure of the discourse.

\paragraph{Step 3: Chunk extraction.}
The argument spans returned by SRL can be long and syntactically 
complex, so each selected span is simplified before being added to 
the set. We remove leading function words, such as determiners and 
prepositions, and extract noun-centered chunks, i.e., shorter spans 
whose head is a common or proper noun, optionally accompanied by 
nearby modifiers. This converts long semantic-role arguments into 
compact entity-like units.

\paragraph{Step 4: Filtering.}
Each extracted chunk $r$ is then checked by two filters. First, 
$\Call{IsInformative}{r}$ removes chunks that do not provide a 
meaningful referential unit. A chunk is retained if it contains 
lexical content headed by a nominal element, such as a common noun, 
proper noun, or named entity, and discarded if it is empty, 
consists only of function words or adverbs, or is purely numeric. 
Numeric tokens are kept when part of a named entity or a 
referential expression. Second, 
$\Call{IsSubjectMention}{r, \mathcal{T}_{e}, e}$ removes chunks 
that correspond to the subject itself, comparing $r$ against $e$, 
its surface variants, and the previously identified subject 
mentions $\mathcal{T}_{e}$.

\paragraph{Step 5: Normalization and merging.}
Finally, the remaining chunks are normalized and merged. Those with 
the same normalized surface form or lemma are treated as repeated 
mentions of the same \reference{} and collapsed into a single 
entry, yielding the \reference{} set $\mathcal{R}_D$.

\begin{algorithm}[t]
\small
\caption{SRL-based \reference{} extraction. Given the subject $e$ 
and a document $D$, the procedure extracts the discourse 
\reference{}s in $D$ while excluding mentions of the subject 
itself.}
\label{alg:srl_reference_extraction}

\begin{algorithmic}[1]
\Require subject $e$, document $D$
\Ensure \reference{} set $\mathcal{R}_D$

\AlgComment{\textbf{Step 1: initialize the set.}}
\State $\mathcal{R}_D \gets \emptyset$
\State $\mathcal{S} \gets \Call{SplitSentences}{D}$

\AlgComment{\textbf{Step 2: identify subject mentions to be excluded later.}}
\State $\mathcal{T}_{e} \gets \Call{FindSubjectMentions}{e, D}$
\Comment{exact matches, variants, anchored partial mentions}

\AlgComment{\textbf{Step 3: extract semantic-role frames from each sentence.}}
\ForAll{$s \in \mathcal{S}$}
    \State $\mathcal{F}_s \gets \Call{SemanticRoleLabeling}{s}$
    \Comment{predicates with labeled argument spans}

    \AlgComment{\textbf{Step 4: keep only semantically relevant argument spans.}}
    \ForAll{$a \in \Call{ArgumentSpans}{\mathcal{F}_s}$}

        \If{\textbf{not} $\Call{IsSelectedRole}{a}$}
            \State \textbf{continue}
        \EndIf

        \State $\mathcal{C} \gets \Call{ExtractChunks}{a}$
        \Comment{trim function words and keep noun-centered chunks}

        \AlgComment{\textbf{Step 5: filter invalid chunks.}}
        \ForAll{$r \in \mathcal{C}$}

            \If{\textbf{not} $\Call{IsInformative}{r}$}
                \State \textbf{continue}
            \EndIf

            \If{$\Call{IsSubjectMention}{r, \mathcal{T}_{e}, e}$}
                \State \textbf{continue}
            \EndIf

            \State $\mathcal{R}_D \gets \mathcal{R}_D \cup \{r\}$
        \EndFor
    \EndFor
\EndFor

\AlgComment{\textbf{Step 6: merge repeated mentions of the same \reference{}.}}
\State $\mathcal{R}_D \gets \Call{NormalizeAndMerge}{\mathcal{R}_D}$
\Comment{same normalized surface or lemma form}

\State \Return $\mathcal{R}_D$

\end{algorithmic}
\end{algorithm}

\subsection{Other implementation details}
Following \citet{kynkaanniemi2019improved}, we set $k=3$ in all
experiments, where $k$ corresponds to the number of nearest neighbors
used to define the radius of each hypersphere. This choice is also
consistent with prior work estimating empirical support as a union of
$k$-nearest-neighbor balls in a PCA-reduced embedding space
\citep{le-bronnec-etal-2024-exploring}. They showed that increasing $k$ enlarges the estimated support and
drives precision and recall toward one, making the overlap criterion
more permissive. We therefore use a small value of $k$ as a conservative
local estimate of support, preserving fine-grained distinctions between
Meaning--Reference configurations and limiting support inflation. We report a quantitative analysis of model rankings across values of 
$k$ in Appendix~\ref{app:k_analysis}. %

The implementation of Semiotic Embeddings relies on a few numerical 
safeguards for stable computation. Before computing 
$\theta_j = \arccos(\langle \widehat{\mathbf{H}}^{(\mu)}, 
\widehat{\mathbf{H}}^{(\rho)}_j \rangle)$, we clip the inner product 
to $[-1,1]$, as commonly done in numerical implementations of 
geometric operations on manifolds \citep{miolane2020geomstats}, so 
that $\theta_j \in [0,\pi]$ by definition. For $0<\theta_j<\pi$, 
Eq.~\ref{eq:semiotic_displacement} gives the closed-form logarithmic 
map. At the two boundary angles, the formula is undefined, so we 
handle them separately. When $\theta_j=0$, the Meaning and Reference 
Embeddings coincide and the displacement is set to the zero vector. 
When $\theta_j=\pi$, the two points are antipodal and are connected by infinitely many shortest geodesics, so we again set the 
displacement to zero instead of choosing an arbitrary tangent 
direction. This convention keeps the construction deterministic and  prevents singular values from propagating to the final Semiotic Embedding.

\subsection{LLM generation prompts}

Figures~\ref{fig:wikipedia_lead_prompt} and~\ref{fig:peircean_semiotics_prompt} 
show the prompts used to query the models in experiments \rone{} and \rtwo{}, 
respectively. In the Wikipedia lead-generation task, the model was provided 
with information about the outputs produced in previous generations. Since the 
terms were extracted from Wikipedia, they can be expected, in many cases, to 
correspond to entities or senses that are relatively salient in the model's 
parametric knowledge. The generation history was therefore included to prevent 
the model from simply restating a previously generated {lead} and, when 
possible, to encourage it to produce a lead for a different sense or for a 
complementary aspect of the same term.

By contrast, no generation history was included in the Peircean semiotics task. In this case, several terms are specialized theoretical concepts for which a single unambiguous definition is not always available. Moreover, Peirce's own terminology and conceptual distinctions were developed and sometimes reformulated across different writings. %

\begin{figure}[t!]
    \centering
    \scalebox{0.700}{
        \begin{tcolorbox}[colback=blue!5!white,colframe=blue!75!black, width=94mm, title=\wikipoly{} prompt]
        
        \textbf{Task-generation:} \\
        {\itshape
        You are a creative writing assistant. Your task is to write a lead section for a given term in a Wikipedia-style encyclopedic, factual, neutral, third-person voice.\\
        The lead may be written as a single paragraph or as multiple short paragraphs when appropriate.\\
        The term may have multiple senses and may refer to common nouns, places, people, organizations, products or brands, and creative works.\\
        Describe exactly one valid sense of the term that is real and attested in public usage or reliable sources.\\
        This request is part of a multi-generation run for the same term.\\
        In this generation, keep the lead strictly about one sense across all paragraphs, which may cover complementary aspects.\\
        If this is generation 1, you are free to choose any one valid sense.\\
        If this is generation 2 or higher, prefer a valid sense that is semantically different from those used in previous generations, not just different wording, when such a sense is available.\\
        The user prompt may include lines in the form ``First 10 words of Gen k: ...'', where k is the generation number of a previous output for the same term.\\
        Use those lines as guidance to infer which senses were already covered and prefer a different valid sense when available, rather than merely paraphrasing an earlier one. %
        }
        
        \noindent \dotfill
        
        \textbf{User prompt:} \\
        {\itshape
        Write a factual, Wikipedia-style lead section for the following term: \textbf{\{term\}}.\\
        This is generation \textbf{\{generation\_number\}} of \textbf{\{generation\_count\}} for this term.\\
        \textbf{\{generation\_history\_block\}}\\
        Prefer a sense that is different from previously used senses for this term, when a distinct valid sense is available.\\
        If not, you may describe the term using the valid sense you judge most appropriate, focusing on a complementary aspect not already covered in previous generations.\\
        Your response must not exceed \textbf{\{max\_words\}} words.\\
        Return only the lead section text, with no preamble and no closing remarks.\\
        Do not invent senses, entities, events, or facts. Do not mention other possible senses.\\
        Do not include titles, headings, comments, bullet points, or explanations.
        }
        
        \end{tcolorbox}
    }
    
\caption{Prompt used in \rone{} for generating a Wikipedia-style lead for a target expression $e$.}
    \label{fig:wikipedia_lead_prompt}
\end{figure}

\begin{figure}[t!]
    \centering
    \scalebox{0.700}{
        \begin{tcolorbox}[colback=blue!5!white,colframe=blue!75!black, width=94mm, title=Peircean concepts prompt]
        
        \textbf{System prompt:} \\
        {\itshape
        You are an expert assistant in Peircean semiotics and academic writing.\\
        Your task is to write a formal and precise encyclopedic entry on a semiotic concept provided as input.\\
        Use domain-specific terminology accurately, and avoid speculative or conversational language.
        }
        
        \noindent \dotfill
        
        \textbf{User prompt:} \\
        {\itshape
        Write an encyclopedic entry for the following concept: \textbf{\{term\}}.\\
        Your response must not exceed \textbf{\{max\_words\}} words.\\
        Return only the entry text, with no preamble and no closing remarks.\\
        Do not include titles, headings, comments, bullet points, or explanations.
        }
        
        \end{tcolorbox}
    }
    
\caption{Prompt used in \rtwo{} for generating definitions of Peircean concepts.}
    \label{fig:peircean_semiotics_prompt}
\end{figure}

\section{Datasets}\label{app:data_details}
Table~\ref{tab:dataset_sizes} reports the size of each dataset used 
in our evaluations. For \rone{} and \rtwo{}, which involve LLM 
generation, each instance was generated $10$ times per temperature, 
with the temperature varied from $0.2$ to $1.6$ in increments of 
$0.2$. 

All datasets are in English and used only for aggregate evaluation.

In the following, we describe the construction of \wikipoly{} and the use of \commens{}, both relying on publicly available textual sources used only for aggregate evaluation. %

\begin{table}[!t]
\centering
\small
\begin{tabular}{@{}llr@{}}
\toprule
Dataset & Comparison unit & Size \\
\midrule
\wikipoly{} & abstract & 936 \\
\peircelc{} & document & 1{,}974 \\
\swipe{} & article & 3{,}279 \\
\oseAI{} & article & 187 \\
\oseAE{} & article& 187 \\
\oseIE{} & article & 187 \\
\commens{} & Peircean concept & 50 \\
\bottomrule
\end{tabular}
\caption{Sizes of the datasets used in our evaluations. The 
comparison unit is the textual element compared between the two 
agents, and the size is the number of such units.}
\label{tab:dataset_sizes}
\end{table}

\subsection{Commens}

The \commens{} \cite{commens} materials are released under a Creative Commons Attribution–NonCommercial–ShareAlike license and are used here for non-commercial research and evaluation purposes only. 

Table~\ref{tab:peirce-top-concepts} lists the Peircean concepts used to query the LLMs and to construct the subset of concepts employed in \rtwo{}. For \rthree{}, we collected all publicly available definitions from \url{http://commens.org/home}.

\begin{table}[htbp]
\centering
\small
\begin{tabular}{@{}p{0.47\linewidth}p{0.47\linewidth}@{}}
\hline
\textbf{Concept} & \textbf{Concept} \\
\hline
1. sign & 2. induction \\
3. symbol & 4. index \\
5. logic & 6. abduction \\
7. interpretant & 8. habit \\
9. real & 10. icon \\
11. pragmatism & 12. deduction \\
13. belief & 14. retroduction \\
15. hypothesis-[reasoning] & 16. categories \\
17. mathematics & 18. existence \\
19. firstness & 20. secondness \\
21. thirdness & 22. object \\
23. rhema & 24. maxim-of-pragmatism \\
25. argument & 26. experience \\
27. proposition & 28. speculative-grammar \\
29. truth & 30. immediate-object \\
31. assertion & 32. methodeutic \\
33. representamen & 34. philosophy \\
35. semeiotic & 36. critic \\
37. metaphysics & 38. reasoning \\
39. science & 40. inference \\
41. logic-[narrow-sense] & 42. percept \\
43. pragmaticism & 44. thought \\
45. continuum & 46. normative-science \\
47. doubt & 48. information \\
49. phaneron & 50. phenomenology \\
\hline
\end{tabular}
\caption{Top Peircean concepts used in this work, selected by the number of definitions available in \commens{} \cite{commens}.}
\label{tab:peirce-top-concepts}
\end{table}

\subsection{\textsc{WikiPoly}}
\label{sec:wikipoly}

We constructed \wikipoly{} from Wikipedia text, available under the Creative Commons Attribution–ShareAlike 4.0 International License (CC BY-SA 4.0) and subject to the Wikimedia Foundation Terms of Use, used here for research and evaluation purposes only. 

\wikipoly{} is a dataset designed to evaluate semantic coverage over highly polysemous topics, where each concept is paired with three to six sense-specific Wikipedia articles sampled from disambiguation pages, favoring candidates with the highest number of distinct senses to avoid bias toward the most popular ones. The result is a controlled collection of polysemous concepts with their principal senses, suitable for evaluating whether a system covers the full semantic range of a concept rather than defaulting to its most common interpretation.

\paragraph{Controlled Diagnostic.}
The controlled diagnostic in Section~\ref{sec:controlled_diagnostic} 
uses \pd{}, an anchor dataset derived from \wikipoly{}, and four 
variants, \qoned{}, \qtwod{}, \qthreed{}, and \qfourd{}. The variants introduce controlled changes in sense coverage and surface 
form. %

\qoned{} applies a semantic perturbation while preserving part of the 
structure of \pd. One paragraph is replaced by another from the 
same Wikipedia sense, leaving {that sense} 
unchanged while altering its textual realization, and a second 
paragraph is replaced by one from a sense of the same term not 
present in \pd. It thus combines a within-sense substitution with 
the introduction of a new sense, testing whether a metric 
distinguishes a surface replacement within one meaning from a 
change in the semantic coverage of the term.

\qtwod{} extends \pd{} by adding further senses of the same term, 
increasing the available semantic material instead of replacing 
existing content. It is therefore broader than \pd, covering more 
{senses} but also introducing content absent from the 
anchor, which tests whether a metric rewards higher coverage or 
penalizes meanings that go beyond the reference.

\qthreed{} is a version of \pd{} retaining only a core subset 
of the original paragraphs and removing part of the semantic 
coverage of the anchor. It remains related to \pd{} but provides a 
less complete representation of the term, testing whether a metric 
is sensitive to missing content and penalizes a dataset that 
preserves some relevant information yet does not cover the full 
range of meanings in \pd.

\qfourd{} preserves the semantic structure of \pd{} while changing the 
surface form of the text, rewriting its paragraphs through lexical 
noise or paraphrastic variation without altering the represented senses 
or referents. It tests robustness to surface-level variation, 
since a metric relying on lexical overlap may assign it a lower 
score, whereas a metric that captures meaning beyond surface form 
should recognize that \qfourd{} retains the same semantic 
organization as the anchor.

\section{Additional experiments}\label{app:additional_experiments}
\subsection{More results on \rtwo}\label{app:rqtwo}
Figures \ref{fig:heatmap_sf} and \ref{fig:heatmap_sc} show \fidelity{} and \coverage{} achieved in the \rtwo{} setting. 

\begin{figure*}[t]
    \centering
    \includegraphics[width=\textwidth]{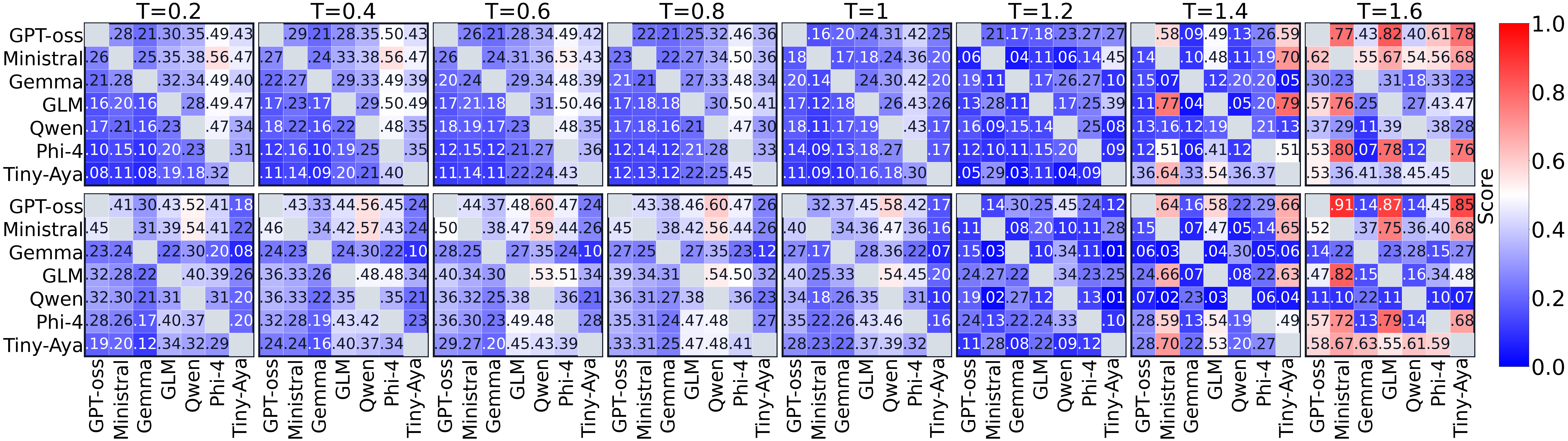}
\caption{\textbf{\rtwo{}}: \ifid{} across temperatures ($T$). Rows and columns correspond to anchors and evaluated models, respectively; 
the shaded diagonal corresponds to 1.}
    \label{fig:heatmap_sf}
\end{figure*}

\begin{figure*}[t]
    \centering
    \includegraphics[width=\textwidth]{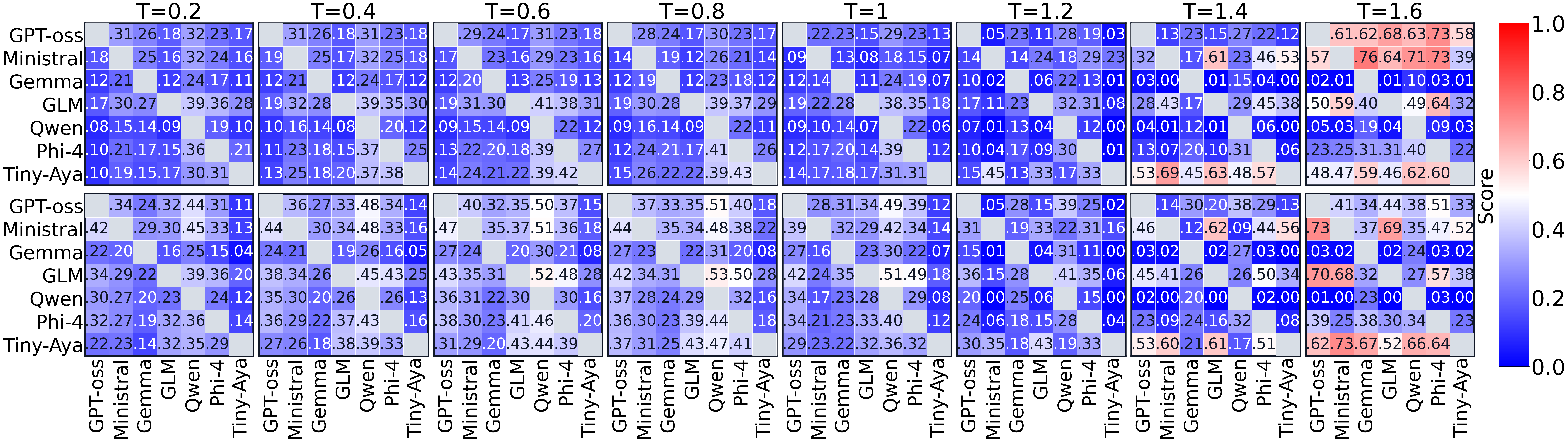}
\caption{\textbf{\rtwo{}}: \icov{} across temperatures ($T$). Rows and columns correspond to anchors and evaluated models, respectively; 
the shaded diagonal corresponds to 1.}
    \label{fig:heatmap_sc}
\end{figure*}

{
\subsection{Results with other MLMs}\label{app:othermlms}

{To assess the sensitivity of our framework across different encoders, we computed \ifid{} and \icov{} on R3, where both the anchor and evaluated texts are human-authored. The original experiments used ModernBERT~\citep{modernbert}, and we repeated the analysis with four additional encoders with MLM-pretrained backbones: BERT~\citep{devlin-etal-2019-bert}, RoBERTa~\citep{liu2019roberta}, DeBERTa~\citep{he2021deberta}, and Jina-ColBERT-v2~\citep{xiao-etal-2024-jina}. Table~\ref{tab:encoder-sensitivity} reports the resulting scores and their harmonic mean for each encoder and dataset.}

\begin{table}[t]
\centering

\resizebox{\columnwidth}{!}{%
\begin{tabular}{@{}lccccc@{}}
\toprule
Dataset & ModernBERT & BERT & RoBERTa & DeBERTa &
\shortstack{Jina-\\ColBERT-v2} \\
\midrule

\multicolumn{6}{c}{\textbf{\fidelity{} (\ifid)}} \\
\midrule
\peircelc{} & .27 & .23 & .24 & .27 & .24 \\
\swipe{}  & .80 & .81 & .80 & .82 & .82 \\
\oseAE{} & .77 & .76 & .75 & .78 & .77 \\
\oseAI{}{} & .91 & .90 & .90 & .92 & .91 \\
\oseIE{} & .85 & .83 & .84 & .86 & .86 \\
\midrule

\multicolumn{6}{c}{\textbf{\coverage{} (\icov)}} \\
\midrule
\peircelc{} & .22 & .23 & .22 & .22 & .20 \\
\swipe{}  & .70 & .71 & .70 & .73 & .74 \\
\oseAE{} & .61 & .59 & .58 & .60 & .60 \\
\oseAI{} & .82 & .81 & .82 & .84 & .82 \\
\oseIE{} & .75 & .74 & .74 & .75 & .74 \\
\midrule

\multicolumn{6}{c}{\textbf{Harmonic Mean}} \\
\midrule
\peircelc{} & .24 & .23 & .23 & .24 & .22 \\
\swipe{}  & .75 & .76 & .75 & .77 & .78 \\
\oseAE{} & .68 & .66 & .66 & .68 & .67 \\
\oseAI{} & .86 & .86 & .86 & .88 & .86 \\
\oseIE{}& .80 & .78 & .79 & .80 & .80 \\
\bottomrule
\end{tabular}%
}
\caption{Sensitivity to different MLM backbones.}
\label{tab:encoder-sensitivity}
\end{table}

{Notably, across all encoders, \ifid{}, \icov{}, and their harmonic mean yield the same ordering of the datasets: OSE-AI $>$ OSE-IE $>$ SWIPE $>$ OSE-AE $>$ PEIRCE. The harmonic mean varies only slightly across encoders within each dataset, suggesting that the overall pattern is consistent across different representation spaces.}
}

{
\subsection{Analysis of Reference Extraction}\label{app:anal_reference_extractor}

\paragraph{Effect of the Reference Extractor.}
To examine the effect of the Reference extractor, we repeated the 
\rthree{} experiments by replacing the SRL-based procedure described 
in Appendix~\ref{app:srl} with an alternative based on noun phrases 
(NPs) and named entity recognition (NER), hereafter \emph{NP+NER}. 
The SRL procedure identifies References from semantically relevant 
argument spans in predicate--argument structures, whereas NP+NER 
directly collects noun chunks and named entities irrespective of their 
semantic roles. We implemented NP+NER with spaCy 
3.8.11\footnote{\url{https://spacy.io/models/en\#en_core_web_sm}} and 
the \texttt{en\_core\_web\_sm} 3.8.0 model. Duplicate spans are removed, 
and the same exclusion criteria as in the SRL procedure are applied. 
Thus, only Reference selection was changed. Meaning and Reference Embeddings were computed with ModernBERT 
in both conditions, and all subsequent representation and scoring steps 
remained unchanged.  Both configurations were evaluated over 20 random seeds, and we report 
the resulting average scores.
We denote the resulting scores by 
\ifid{}$_{\mathrm{NP+NER}}$ and \icov{}$_{\mathrm{NP+NER}}$.

\begin{table}[t]
\centering
\footnotesize
\setlength{\tabcolsep}{3pt}
\renewcommand{\arraystretch}{0.95}
\begin{tabular*}{\columnwidth}{@{\extracolsep{\fill}}lcccc@{}}
\toprule
Dataset & \ifid{} & \icov{} & 
\ifid{}$_{\mathrm{NP+NER}}$ & \icov{}$_{\mathrm{NP+NER}}$ \\
\midrule
\peircelc{} & .27 & .22 & .17 & .14 \\
\swipe{}    & .80 & .70 & .76 & .67 \\
\oseAE{}    & .77 & .61 & .75 & .59 \\
\oseAI{}    & .91 & .82 & .90 & .82 \\
\oseIE{}    & .85 & .75 & .84 & .74 \\
\bottomrule
\end{tabular*}
\caption{\ifid{} and \icov{} in \rthree{} using the SRL-based and  NP+NER Reference extractors.}
\label{tab:reference-extractor}
\end{table}

Using either extractor, the fidelity and coverage scores yielded the 
same ordering: \oseAI{} $>$ \oseIE{} $>$ \swipe{} $>$ \oseAE{} $>$ 
\peircelc{}. For each measure, Kendall's $\tau_b=1.0$ indicates 
agreement across every pairwise comparison. The NP+NER extractor generally produced lower absolute scores, with the largest difference on 
\peircelc{} and smaller differences across the \ose{} comparisons. 
Thus, the extraction algorithm affects the absolute values while 
preserving the same comparative pattern in \rthree{} across the two 
tested approaches.
\paragraph{Effect of Reference Omission.}
\label{app:reference_omission}
To examine the effect of missed References on \ifid{} and \icov{}, 
we simulated their omission in the \rthree{} experiments shown in 
Fig.~\ref{fig:hvh}. We independently removed $10\%$, $20\%$, 
and $30\%$ of the extracted Reference points from the anchor and 
evaluated collections without replacement over 20 random seeds, while 
retaining all Meaning points. For each condition, we recomputed the PCA 
projection, the $k$-NN supports with $k=3$, \ifid{}, and \icov{}. 
We then computed the absolute differences between the resulting scores 
and those obtained with all extracted Reference points retained.

\begin{table}[t]
\centering
\footnotesize
\renewcommand{\arraystretch}{0.95}
\begin{tabular*}{\columnwidth}{@{\extracolsep{\fill}}lccc@{}}
\toprule
Dataset & $10\%$ & $20\%$ & $30\%$ \\
\midrule
\multicolumn{4}{c}{\textbf{\fidelity{} (\ifid{})}} \\
\midrule
\peircelc{} & .02 & .04 & .06 \\
\swipe{}    & .01 & .02 & .03 \\
\oseAE{}    & .01 & .03 & .05 \\
\oseAI{}    & .02 & .04 & .06 \\
\oseIE{}    & .02 & .04 & .06 \\
\midrule
\multicolumn{4}{c}{\textbf{\coverage{} (\icov{})}} \\
\midrule
\peircelc{} & .01 & .03 & .04 \\
\swipe{}    & .01 & .03 & .04 \\
\oseAE{}    & .02 & .03 & .05 \\
\oseAI{}    & .02 & .04 & .07 \\
\oseIE{}    & .02 & .04 & .06 \\
\bottomrule
\end{tabular*}
\caption{Mean absolute differences in \ifid{} and \icov{} in 
\rthree{} after removing $10\%$, $20\%$, or $30\%$ of the Reference 
points. Differences are computed against scores obtained with all 
extracted Reference points retained.}
\label{tab:reference-omission}
\end{table}

Across all removal rates, \ifid{} and \icov{} yielded the same 
ordering: \oseAI{} $>$ \oseIE{} $>$ \swipe{} $>$ \oseAE{} $>$ 
\peircelc{}. Additionally, Kendall's $\tau_b=1.0$ indicates complete 
agreement with the baseline ordering. The largest mean absolute changes 
were $.02$ and $.02$ at $10\%$, $.04$ and $.04$ at $20\%$, and $.06$ 
and $.07$ at $30\%$, for \ifid{} and \icov{}, respectively. Overall, 
the scores responded progressively to Reference omission while 
preserving the comparative conclusions in \rthree{} across all tested 
conditions.

}

\subsection{Analysis of $k$}\label{app:k_analysis}
Figure~\ref{fig:k_analysis} shows, for each model, how \fidelity{} 
and \coverage{} vary with $k$, averaged over the available 
temperatures. %
Both scores increase with $k$ for every model, 
as expected from support-based estimators, where larger 
neighborhoods yield larger hyperspheres and thus higher overlap 
\citep{le-bronnec-etal-2024-exploring}.
{We also examined how model rankings change with $k$ in 
order to quantify the effect of neighborhood scale on the differences 
observed among models. For each of the nine alternative values 
$k\in\{1,2,4,5,6,7,8,9,10\}$, we compared the ranking of the seven 
models with that at $k=3$, separately for \ifid{} and \icov{}, using 
Kendall's $\tau_b$. Combining the nine alternative values of $k$ with 
the two measures yielded $9\times2=18$ ranking comparisons, comprising 
378 pairwise model orderings. Of these, 357 (94.4\%) were preserved. 
The minimum $\tau_b$ was $.810$ for \ifid{} and $.714$ for \icov{}, 
with 17 comparisons yielding at least $.810$ and no model shifting by 
more than two positions. For the inverted pairs, the median (maximum) 
absolute gaps at $k=3$ were $.010$ ($.033$) and $.013$ ($.025$), while 
at the values of $k$ where the ordering reversed they were $.012$ 
($.028$) and $.010$ ($.018$), for \ifid{} and \icov{}, respectively. 
Thus, the observed ranking changes tended to concern models separated 
by relatively small score margins.}

\begin{figure*}[t]
    \centering

    \begin{minipage}{0.48\textwidth}
        \centering
        \includegraphics[width=\linewidth]{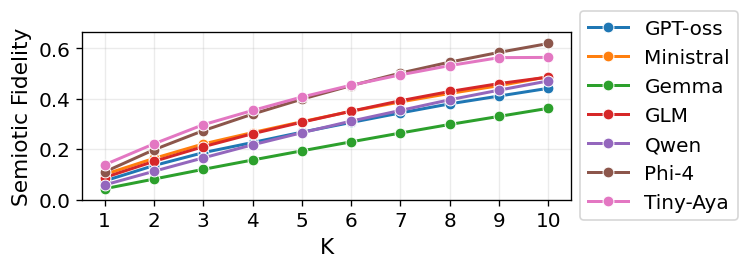}
    \end{minipage}
    \hfill
    \begin{minipage}{0.48\textwidth}
        \centering
        \includegraphics[width=\linewidth]{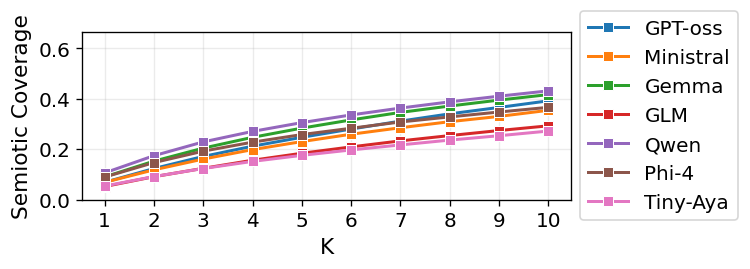}
    \end{minipage}

\caption{\fidelity{} (\ifid)  (left) and \coverage{} (\icov)  (right) across values of $k$, averaged 
over temperatures, on \wikipoly{} in the \rone{} setting.}
    \label{fig:k_analysis}
\end{figure*}

\section{Monotonicity of \coverage{}}\label{app:monotonicity}

Here, we provide a formal proof of the monotonicity property of 
\coverage{}. For a fixed \expression{} $e$, let 
$\mathcal{Z}^{(\alpha^{*})}_{e}$ and $\mathcal{Z}^{(\beta)}_{e}$ 
denote the anchor and evaluated semiotic collections, 
respectively. For each $\mathbf{x}\in\mathcal{Z}^{(\beta)}_{e}$, 
recall from Eq.~\ref{eq:semiotic_fidelity_coverage} that the 
anchor-calibrated radius is defined as:
\begin{equation}
\begin{aligned}
&r_k\!\left(\pi(\mathbf{x}), \mathcal{Z}^{(\alpha^{*})}_{e}\right), \\
&\pi(\mathbf{x}) =
\arg\min_{\mathbf{a}\in \mathcal{Z}^{(\alpha^{*})}_{e}}
\|\mathbf{x}-\mathbf{a}\|_2,
\end{aligned}
\end{equation}
\noindent
with ties resolved by any fixed rule. In our implementation, when 
two or more anchor points achieve the minimum distance to 
$\mathbf{x}$, we select the first one in enumeration order. The 
support induced by $\mathcal{Z}^{(\beta)}_{e}$ at the anchor scale 
is
\begin{equation}
\begin{aligned}
\mathcal{S}\!\left(\mathcal{Z}^{(\beta)}_{e}\right)
&=
\bigcup_{\mathbf{x}\in\mathcal{Z}^{(\beta)}_{e}}
\left\{\mathbf{y}: \right. \\
&\left.
\|\mathbf{y}-\mathbf{x}\|_2 \le 
r_k\!\left(\pi(\mathbf{x}), \mathcal{Z}^{(\alpha^{*})}_{e}\right)
\right\},
\end{aligned}
\end{equation}
so that Eq.~\ref{eq:semiotic_fidelity_coverage} can be rewritten as
\begin{equation*}
\icov_e(\beta \mid \alpha^{*})
=
\frac{1}{|\mathcal{Z}^{(\alpha^{*})}_{e}|}
\sum_{\mathbf{a}\in \mathcal{Z}^{(\alpha^{*})}_{e}}
\mathbb{I}\!\left[\mathbf{a}\in 
\mathcal{S}\!\left(\mathcal{Z}^{(\beta)}_{e}\right)\right].
\end{equation*}
Let $\mathcal{Z}^{(\beta')}_{e}\subseteq\mathcal{Z}^{(\beta)}_{e}$ 
be obtained by removing one or more semiotic points from the 
evaluated collection. Since the radius 
$r_k(\pi(\mathbf{x}), \mathcal{Z}^{(\alpha^{*})}_{e})$ depends only 
on $\mathbf{x}$ and on the fixed anchor collection 
$\mathcal{Z}^{(\alpha^{*})}_{e}$, the hyperspheres associated with 
the remaining points are unchanged. Therefore
\begin{equation*}
\mathcal{S}\!\left(\mathcal{Z}^{(\beta')}_{e}\right)
\subseteq
\mathcal{S}\!\left(\mathcal{Z}^{(\beta)}_{e}\right),
\end{equation*}
and for every anchor point 
$\mathbf{a}\in\mathcal{Z}^{(\alpha^{*})}_{e}$,
\begin{equation*}
\mathbb{I}\!\left[\mathbf{a}\in 
\mathcal{S}\!\left(\mathcal{Z}^{(\beta')}_{e}\right)\right]
\le
\mathbb{I}\!\left[\mathbf{a}\in 
\mathcal{S}\!\left(\mathcal{Z}^{(\beta)}_{e}\right)\right].
\end{equation*}
Averaging the inequality over all anchor points yields:
\begin{equation*}
\begin{aligned}
\frac{1}{|\mathcal{Z}^{(\alpha^{*})}_{e}|}
\sum_{\mathbf{a}\in \mathcal{Z}^{(\alpha^{*})}_{e}}
\mathbb{I}\!\left[\mathbf{a}\in 
\mathcal{S}\!\left(\mathcal{Z}^{(\beta')}_{e}\right)\right] 
&\le \\
\frac{1}{|\mathcal{Z}^{(\alpha^{*})}_{e}|}
\sum_{\mathbf{a}\in \mathcal{Z}^{(\alpha^{*})}_{e}}
\mathbb{I}\!\left[\mathbf{a}\in 
\mathcal{S}\!\left(\mathcal{Z}^{(\beta)}_{e}\right)\right]&,
\end{aligned}
\end{equation*}
which, by the rewriting above, is equivalent to
\begin{equation*}
\icov_e(\beta' \mid \alpha^{*})
\le
\icov_e(\beta \mid \alpha^{*}).
\end{equation*}
Thus, removing points from $\mathcal{Z}^{(\beta)}_{e}$ cannot 
increase \icov{}, since the radii are calibrated only on 
$\mathcal{Z}^{(\alpha^{*})}_{e}$ and the support 
$\mathcal{S}\!\left(\mathcal{Z}^{(\beta)}_{e}\right)$ can therefore 
only shrink under sparsification. \hfill$\square$

\section{Additional Background}
\label{app:additional_background}

\subsection{More on Semiotics}

Semiotics studies signs and the processes through which forms acquire meaning for interpreting agents. A visible, audible, or written form functions as a sign only when taken to stand for something within a context of interpretation \citep{peirce1931}, establishing a relation between signs, what they are about, and the interpretive activity that gives rise to sense.
A classical account is offered by de Saussure, who describes the sign as a dyadic relation between a \emph{signifier}, the perceptible form, and a \emph{signified}, the associated concept \citep{de1989cours}. In this model, language is a structured system of differences where each word derives its value from its position within the system. For our purposes, however, the Saussurean model falls short on two points. First, it does not assign the interpreting agent a structural role, leaving variation across interpreters outside the theory. Second, it does not treat the relation to an object as constitutive of the sign, confining meaning to a relation between two internal poles of the linguistic system. Peirce's framework is better suited to our needs because it casts meaning as an interpretive process linking a sign to what it is about through the effect produced in an agent.

In Peirce's formulation, a sign is a triadic relation among three inseparable elements: the \emph{representamen}, the form through which the sign appears, the \emph{object}, what the sign is about, and the \emph{interpretant}, the understanding, effect, or further sign produced when the representamen is taken to stand for the object \citep{peirce1931}. Meaning emerges from this relation as a whole: a representamen without an object would not be about anything, an object without an interpretant would not be taken as meant by any sign, and an interpretant without a representamen would have nothing to arise from. Semiosis is therefore not a direct act of naming, but a mediated process in which form, object, and interpretive effect constitute a single relation.

Within this structure, the interpretant plays a stabilizing role, anchoring the sign to a specific agent on a given occasion and fixing the respect in which the representamen stands for its object. Without it, the link between form and object would remain underdetermined, since the same form can in principle be related to many different objects. The interpretant closes this gap by grounding one such respect in the understanding of an interpreter, so that semiosis is always performed from the standpoint of an agent who brings intentionality to the sign. This active contribution can be described more precisely with the notion of a \emph{cultural unit} introduced by \citet{eco1984semiotics}. For Eco, the content side of a sign is not a private mental image but a culturally codified segment of the shared semantic system that a community recognizes as a stable meaning. In Peircean terms, what counts as the object of a sign is filtered by the categories, experiences, and practices available to the interpreter, and the interpretant is the locus where these cultural units become operative for an individual agent, turning the shared code into a concrete interpretive act.

Figure~\ref{fig:example} illustrates this point with the representamen \emph{Smoke}. An agent perceives the form \emph{smoke} and produces the interpretant \emph{``There is fire!''}, which makes the sign relation determinate. Through this interpretant, the object is articulated as a structured set of cultural units, including \emph{smoke}, \emph{fire}, and \emph{wood}, acquired from a shared cultural background. A different interpreter, equipped with different cultural units, could relate the same representamen to a different object, for instance to a ritual signal or to an industrial process.

\begin{figure}
    \centering
    \includegraphics[width=1.0\linewidth]{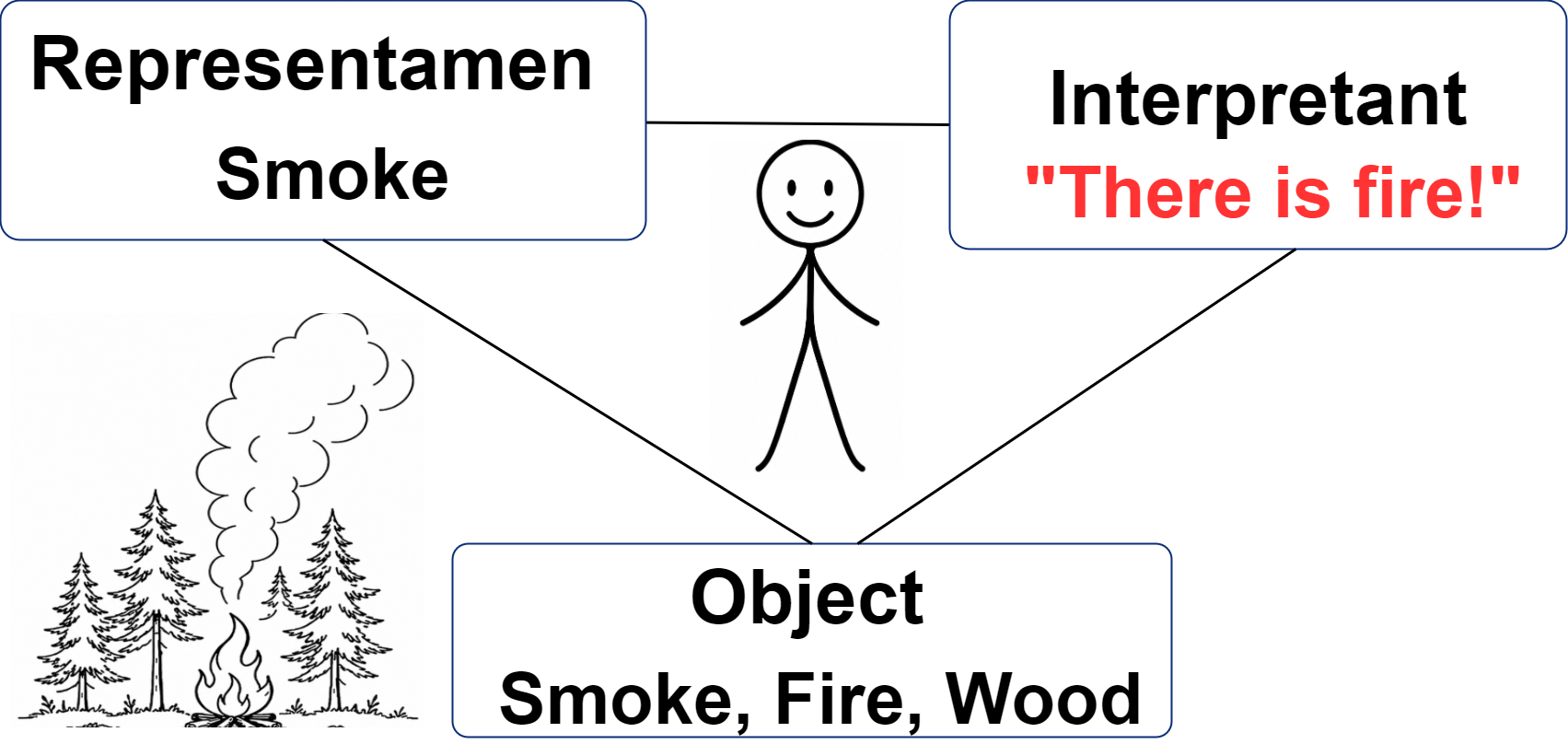}
    \caption{Peircean triadic semiosis for the representamen
    \emph{Smoke}. The interpreter, at the center of the relation,
    produces the interpretant \emph{There is fire!}, which stabilizes
    the sign by relating the representamen to an object articulated as
    a set of cultural units (\emph{smoke}, \emph{fire}, \emph{wood}). The lines indicate the
    inseparable relation among representamen, interpretant, and object.}
    \label{fig:example}
\end{figure}

\subsection{Logarithmic Map}
\label{app:spherical-tangent-displacements}
In this work, we employed the logarithmic map to represent each 
Reference Embedding as a displacement from the Meaning Embedding in 
the local tangent space at the Meaning Embedding itself. This operation
provides a geometry-motivated way to express each Reference 
relative to the Meaning on the normalized embedding sphere following semiotic principles. %

To apply this construction, 
both $\mathbf{H}^{(\mu)}$ and $\mathbf{H}^{(\rho)}_{j}$ are first 
normalized onto the unit hypersphere, yielding 
$\widehat{\mathbf{H}}^{(\mu)}$ and 
$\widehat{\mathbf{H}}^{(\rho)}_{j}$. On the unit hypersphere, a 
Euclidean difference would describe a displacement in the  surrounding Euclidean 
space, whereas the logarithmic map gives a tangent vector anchored 
at the Meaning Embedding, which makes the Reference relative to 
the Meaning, as required by the semiotic model described in Sec. \ref{sec:preliminaries}. Specifically, the angle 
$\theta_j = \operatorname{\arccos}\!\left(\langle 
\widehat{\mathbf{H}}^{(\mu)}, \widehat{\mathbf{H}}^{(\rho)}_{j} 
\rangle \right) \in [0,\pi]$ is the spherical distance between the 
two normalized embeddings. To obtain the tangent direction from the Meaning to the Reference, 
the component of $\widehat{\mathbf{H}}^{(\rho)}_{j}$ parallel to 
$\widehat{\mathbf{H}}^{(\mu)}$ is removed:
\begin{equation}
\widehat{\mathbf{H}}^{(\rho)}_{j}
-
\cos(\theta_j)\widehat{\mathbf{H}}^{(\mu)}.
\end{equation}
\noindent
This vector belongs to the tangent space at the Meaning Embedding, 
since it is orthogonal to $\widehat{\mathbf{H}}^{(\mu)}$:
\begin{equation}
\left\langle
\widehat{\mathbf{H}}^{(\rho)}_{j}
-
\cos(\theta_j)\widehat{\mathbf{H}}^{(\mu)},
\widehat{\mathbf{H}}^{(\mu)}
\right\rangle
=
0.
\end{equation}

Hence, for $0<\theta_j<\pi$, the logarithmic map gives
\begin{equation}
\mathbf{H}^{(\chi)}_{j}
=
\theta_j
\frac{
\widehat{\mathbf{H}}^{(\rho)}_{j}
-
\cos(\theta_j)\,\widehat{\mathbf{H}}^{(\mu)}
}{
\left\|
\widehat{\mathbf{H}}^{(\rho)}_{j}
-
\cos(\theta_j)\,\widehat{\mathbf{H}}^{(\mu)}
\right\|
}.
\end{equation}

The numerator gives the tangent direction at 
$\widehat{\mathbf{H}}^{(\mu)}$ pointing toward 
$\widehat{\mathbf{H}}^{(\rho)}_{j}$, the denominator normalizes 
this direction to unit length, and the factor $\theta_j$ scales it 
by the spherical distance between the two embeddings. The 
resulting vector $\mathbf{H}^{(\chi)}_{j}$ therefore belongs to the 
tangent space at $\widehat{\mathbf{H}}^{(\mu)}$ and encodes the 
Reference Embedding through both the direction and the magnitude 
of its displacement from the Meaning Embedding.

\begin{table*}[t]
\centering
\small
\renewcommand{\arraystretch}{1.12}
\setlength{\tabcolsep}{6pt}
\begin{tabular}{p{0.26\textwidth}p{0.68\textwidth}}
\toprule
\textbf{Symbol} & \textbf{Description} \\
\midrule

\(\mathcal{E}\) & Set of {target expressions under analysis}. \\

\(e \in \mathcal{E}\) & Generic {target expression under analysis}. \\

\(\gamma \in \{\beta,\alpha^{*}\}\) & Generic agent; \(\beta\) is the evaluated agent and \(\alpha^{*}\) is the anchor agent. \\

\(\mathcal{D}^{(\gamma)}_{e}\) & Collection of texts produced by agent \(\gamma\) describing expression \(e\). \\

\(D \in \mathcal{D}^{(\gamma)}_{e}\) & Generic text produced by agent \(\gamma\) for expression \(e\). \\

\(\mathcal{Z}^{(\gamma)}_{e}\) & Set of Semiotic Embeddings associated with agent \(\gamma\) for expression \(e\). \\

\(F\) & Transformer encoder pre-trained with a masked language modeling objective. \\

\(d\) & Hidden dimensionality of the encoder \(F\). \\

\(w_1,\dots,w_n\) & Occurrences of expression \(e\) in document \(D\), where \(n\) is their number. \\

\(w_i\) & The \(i\)-th occurrence of the target expression \(e\). \\

\(c_i\) & Context window associated with occurrence \(w_i\). \\

\(\mathrm{mask}(c_i,w_i)\) & Context window \(c_i\) where \(w_i\) is replaced by the mask token. \\

\(\mathbf{H}^{(\mu)}_i\) & Masked contextual representation of occurrence \(w_i\). \\

\(\mathbf{H}^{(\mu)}\) & Meaning Embedding obtained by averaging all \(\mathbf{H}^{(\mu)}_i\). \\

\(\mathcal{R}_D = \{r_1,\dots,r_m\}\) & Set of references associated with expression \(e\) in document \(D\), where \(m\) is their number. \\

\(r_j \in \mathcal{R}_D\) & Generic reference associated with \(e\). \\

\(w_{j,1},\dots,w_{j,n_j}\) & Occurrences of reference \(r_j\) in document \(D\), where \(n_j\) is their number. \\

\(w_{j,l}\) & The \(l\)-th occurrence of reference \(r_j\). \\

\(c_{j,l}\) & Context window associated with occurrence \(w_{j,l}\). \\

\(\mathbf{H}^{(\rho)}_j\) & Reference Embedding associated with reference \(r_j\). \\

\(\widehat{\mathbf{H}}^{(\mu)}\) & Normalized Meaning Embedding projected onto the unit hypersphere. \\

\(\widehat{\mathbf{H}}^{(\rho)}_j\) & Normalized Reference Embedding projected onto the unit hypersphere. \\

\(\theta_j\) & Angular distance between \(\widehat{\mathbf{H}}^{(\mu)}\) and \(\widehat{\mathbf{H}}^{(\rho)}_j\). \\

\(\langle \cdot,\cdot \rangle\) & Euclidean inner product. \\

\(\|\cdot\|\), \(\|\cdot\|_2\) & Euclidean norm. \\

\(\mathbf{H}^{(\chi)}_j\) & Tangent displacement encoding the reference relative to the Meaning Embedding. \\

\(\mathbf{0}_d\) & Zero vector in \(\mathbb{R}^{d}\). \\

\([\,\cdot\,\Vert\,\cdot\,]\) & Vector concatenation operator. \\

\(\mathbf{H}^{(\sigma)}_j\) & Semiotic Embedding combining Meaning and Reference displacement. \\

\(\mathbb{R}^{d}\), \(\mathbb{R}^{2d}\) & Vector spaces of Meaning/Reference Embeddings and Semiotic Embeddings, respectively. \\

\(\widetilde{\mathbf{H}}^{(\sigma)}_{D,j}\) & PCA-projected Semiotic Embedding in the shared low-dimensional space. \\

\(m_{\gamma,D}\) & Number of references extracted from document \(D\) for agent \(\gamma\). \\

\(k\) & Neighborhood parameter used for \(k\)-nearest-neighbor support estimation. \\

\(\mathbf{N}_k(\widetilde{\mathbf{H}}^{(\sigma)}_{D,j},\mathcal{Z}^{(\gamma)}_e)\) & \(k\)-th nearest neighbor of \(\widetilde{\mathbf{H}}^{(\sigma)}_{D,j}\) in \(\mathcal{Z}^{(\gamma)}_e\). \\

\(r_k(\widetilde{\mathbf{H}}^{(\sigma)}_{D,j},\mathcal{Z}^{(\gamma)}_e)\) & Radius defined by the distance to the \(k\)-th nearest neighbor. \\

\(\mathbf{x} \in \mathcal{Z}^{(\beta)}_e\), \(\mathbf{a} \in \mathcal{Z}^{(\alpha^{*})}_e\) & Generic evaluated and anchor semiotic points, respectively. \\

\(\mathbb{I}[\cdot]\) & Indicator function. \\

\(\pi(\mathbf{x})\) & Nearest anchor point to \(\mathbf{x}\). \\

\(\mathrm{SF}_e(\beta \mid \alpha^{*})\), \(\mathrm{SC}_e(\beta \mid \alpha^{*})\) & Semiotic Fidelity and Semiotic Coverage for expression \(e\). \\

\(\mathrm{SF}(\beta \mid \alpha^{*})\), \(\mathrm{SC}(\beta \mid \alpha^{*})\) & Dataset-level Semiotic Fidelity and Semiotic Coverage averaged over all expressions. \\

\(R1,R2,R3\) & Experimental settings used in the paper. \\

\(T\) & Sampling temperature used during LLM generation. \\

\bottomrule
\end{tabular}
\caption{Summary of the main symbols and notation used throughout the paper.}
\label{tab:nots}
\end{table*}

\end{document}